\documentclass[journal=jacsat,manuscript=article]{achemso}

\usepackage[version=3]{mhchem} 
\usepackage{xcolor}
\usepackage{soul}
\usepackage{gensymb}
\usepackage{graphicx}
\usepackage{hyperref}
\usepackage{pdfpages}

\author{Yayati Jadhav$^1$}
\affiliation{
  Department of Mechanical Engineering, Carnegie Mellon University, Pittsburgh,
  PA, USA
}

\author{Peter Pak}
\affiliation{
  Department of Mechanical Engineering, Carnegie Mellon University, Pittsburgh,
  PA, USA
}

\author{Amir Barati Farimani*}
\email{barati@cmu.edu}
\affiliation{
  Department of Mechanical Engineering, Carnegie Mellon University, Pittsburgh,
  PA, USA
}
\alsoaffiliation{
  Machine Learning Department, Carnegie Mellon University, Pittsburgh, PA, USA
}

\title[]{LLM-3D Print: Large Language Models To Monitor and Control 3D Printing}

\abbreviations{IR,NMR,UV}

\begin{document}

\textbf{Website:} \url{https://sites.google.com/andrew.cmu.edu/printerchat}
\begin{abstract}

Industry 4.0 has revolutionized manufacturing by driving digitization and shifting the paradigm toward additive manufacturing (AM). Material extrusion (MEX), a core AM method, produces customized and cost-effective products with minimal waste, challenging traditional subtractive manufacturing. Despite its advantages, MEX remains susceptible to defects that can compromise part quality and function, often requiring expert intervention. Existing rule-based and machine learning approaches struggle to generalize across different printers and sensors, while deep learning methods depend on large labeled datasets, limiting their scalability and adaptability.
To address these challenges, we introduce a process monitoring and control framework that employs Large Language Models (LLMs) as autonomous controllers for additive manufacturing. Unlike rule-based or heavily data-dependent approaches, our method requires no domain-specific fine-tuning or training. Instead, the LLM leverages in-context learning, self-prompting, and iterative prompt-reason refinement to evaluate print quality from sequential image captures, detect and classify emerging failure modes, and query and modify the printer for relevant operating parameters. Through this adaptive reasoning process, the LLM not only interprets defects but also improves its own decision-making logic, autonomously formulating and executing corrective actions. This demonstrates a rule-free, self-improving approach to process control that extends beyond traditional quality assurance methods. We validated the effectiveness of the proposed framework by comparing it with a control group of engineers with different levels of AM expertise. The evaluation showed that LLM-based agents not only reliably identified common 3D printing errors such as inconsistent extrusion, stringing, warping, and poor layer adhesion, but also determined their causes and corrected them without human intervention. In addition to matching expert-level accuracy, the LLM was able to recognize emerging print errors earlier than human experts, highlighting its value as a proactive controller. To further demonstrate generalizability, we deployed and tested the framework on two different 3D printers with distinct sensor setups, confirming its adaptability across hardware. We also performed compression tests on baseline prints and on prints optimized by the LLM, with the optimized parts showing clear improvements in mechanical performance.

\end{abstract}
\begin{figure}[h]
  \centering
  \includegraphics[width=0.75\textwidth]{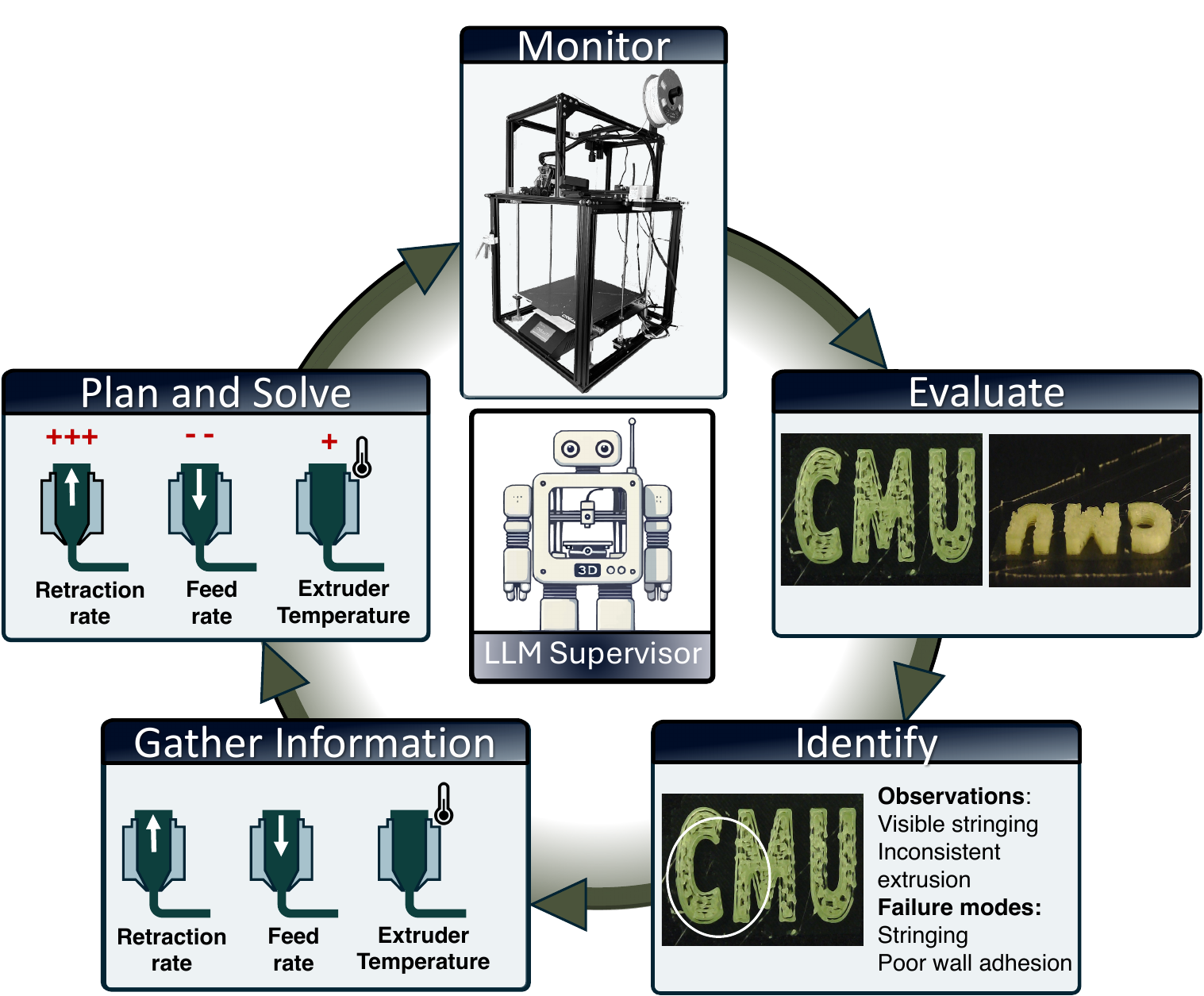}
  \caption*{ \textbf{Abstract Figure: LLMs in continuous improvement cycle} LLM-based supervisor agents can be employed at each step of the continuous improvement cycle. The cycle involves evaluating print quality, identifying failure modes, gathering relevant information, and planning and solving the issues by adjusting the print parameters, ensuring high-quality defect-free parts.}
\end{figure} 

\section{Introduction}

Industry 4.0, also known as the Fourth Industrial Revolution, integrates the
Internet of Things (IoT), artificial intelligence, and big data analytics to
assist in smart manufacturing where machines and systems seamlessly
communicate to optimize production processes \cite{gunal2019simulation, pascoal2023digitalisation}. This technological integration
enhances quality control and certification processes via \textit{in-situ}
process monitoring, utilizing automated data collection to ensure
consistent adherence to quality standards \cite{Bauza2018Realization}. Automated
documentation facilitates comprehensive traceability, simplifies certification,
and provides verifiable records that help manufacturers achieve and maintain
industry certifications, boosting compliance and customer trust
\cite{Wang2019In-Place}.

Additive Manufacturing (AM) plays a key role in this industrial revolution,
enabling rapid prototyping from conceptual design to production level parts
through an iterative design process, bypassing the constraints of conventional
manufacturing processes \cite{yadroitsev2021fundamentals}. AM significantly
reduces production time, enhances design flexibility, lowers costs, minimizes
post-processing, and supports multi-material design through a layer-by-layer
printing process\cite{Jin2020Machine}. This gives AM a substantial advantage over
traditional subtractive manufacturing techniques, such as CNC machining, by
offering greater material efficiency and design versatility
\cite{faludi2015comparing}. Consequently, AM has shown promising applications in
various fields, including healthcare \cite{buj2021use, giannatsis2009additive},
medical devices \cite{da2021comprehensive, haghiashtiani20203d}, and aerospace
\cite{najmon2019review}, among others \cite{stano2021additive,
o2014advances,bacciaglia2020evaluation}.


One of the most widely used additive manufacturing processes is Material Extrusion (MEX). This technique involves extruding thermoplastic filaments through a heated nozzle, which deposits material layer by layer to construct a part from the bottom up \cite{gibson2021additive}. The accessibility and versatility of MEX have fueled the growth of open-source projects like RepRap \cite{sells2010reprap, jones2011reprap} and Klipper\cite{noauthor_klipper3dklipper_2025}. These community driven projects have democratized 3D printing and have enabled individuals, small businesses \cite{laplume2016open}, and research labs \cite{pearce2012building, pearce2013open} to develop and customize hardware, significantly expanding the applications and accessibility of additive manufacturing.

Despite its advantages, MEX is highly susceptible to errors and failures, which limit its consistency and material efficiency. Common issues include warpage, layer misalignment, under-extrusion, over-extrusion, and stringing, among others.
\cite{peng2012investigation, agarwala1996structural, hsiang2020overview, all3dp_3d_printing_problems, nuchitprasitchai2017factors}
These issues are further compounded by the lack of standardized hardware, material variations \cite{Samykano2019Mechanical}, user errors \cite{Song2019Causes}, and print process parameters that are not tailored to specific designs \cite{Dey2019A, Baraheni2023Effects}, all of which contribute to significant variations in print quality. Studies indicate that 20\% of PLA prints fail \cite{wittbrodt2013life}, ABS prints have a material waste rate of 34\% \cite{song2017material}, and overall failure rates can reach 41.1\%, with human error accounting for 26.3\% of these failures \cite{Song2019Causes}. These failures not only result in wasted material, energy, and time but also limit the use of AM parts in end-use products, especially in safety-critical applications like medical devices and aerospace components.

Some approaches address print failures by simulating the 3D printing process and optimizing part-specific parameters \cite{Garzon-Hernandez2020Design, Dey2019A}. While effective, these simulations are computationally expensive and impractical for production, especially when compared to the lower inference cost/time of LLMs, which operate on a per-layer basis in seconds rather than hours. Another strategy targets the limitations of open-loop MEX systems by introducing closed-loop control, where “real-time” refers to parameter adjustments made at the end of each print layer or even a more finer sampling rate. Prior work has incorporated sensors such as accelerometers \cite{li2019prediction}, acoustic sensors \cite{liu2018improved}, and laser scanners \cite{faes2016process} to detect anomalies. However, many defects, such as stringing or subtle layer separation, lack strong physical signatures and remain difficult to detect with such sensors. Additionally, the cost and complexity of these sensor systems hinder their widespread use \cite{brion2022generalisable}, motivating more generalizable, multimodal approaches.

In contrast, camera-based monitoring systems offer a low-cost and easy-to-integrate alternative for identifying surface defects and other visible issues in 3D printing \cite{baumann2016vision, moretti2020towards, cheng2008vision, he2019profile}. Infrared or thermal cameras can detect anomalies not visible to standard visual cameras \cite{costa2017estimation, dinwiddie2013real}, and multi-camera setups can reconstruct 3D models of printed parts to check dimensional and shape accuracy \cite{holzmond2017situ, fastowicz2019objective}. Due to their cost-effectiveness and simplicity, standard camera systems have seen more widespread adoption. A notable study developed an image-based closed-loop quality control system for fused filament fabrication, implemented by a customized online image acquisition system with a proposed image diagnosis-based feedback quality control method \cite{Liu_Law_Roberson_Kong_2019}. While computer vision approaches are promising for targeting specific errors, they often require calibration for each part, printer, and material, making it difficult to create feature extraction algorithms that generalize across different setups. Consequently, these methods typically work with only a single combination of printer, part geometry, material, and printing conditions. Integrating machine learning techniques offers a more flexible and adaptable solution, enabling robust error detection and correction across diverse printing environments.

Machine learning, particularly deep learning techniques, has achieved state-of-the-art performance in various science and engineering applications. These techniques have been successfully applied to surrogate modeling \cite{Jadhav_Berthel_Hu_Panat_Beuth_Farimani_2023,ogoke2023inexpensive,ogoke2024deep, strayer2022accelerating}, generating 3D printable structures \cite{Jadhav_Berthel_Hu_Panat_Beuth_Farimani_2024}, and predicting failures in 3D printing \cite{pak2024thermopore,Tian2020Deep,taherkhani2022unsupervised}, among other areas \cite{caomachine,ajenifujah2024integrating}. These advancements demonstrate the significant potential of machine learning to enhance and innovate within the field of additive manufacturing.

Recent research has leveraged machine learning for error detection in MEX, demonstrating promising results through camera-based monitoring and convolutional neural networks to predict and correct issues such as extrusion rate \cite{Jin_Zhang_Gu_2019}, warpage \cite{saluja2020closed}, surface defects \cite{delli2018automated}, and layer defects \cite{jin2020automated}. One notable approach aims to generalize the process across different parts, materials, and printing systems by using convolutional neural networks to detect errors and adjust parameters such as flow rate, speed, z-offset, and extruder temperature \cite{Brion_Pattinson_2022}.

However, most of these methods are limited to addressing errors within a single modality by changing only a few print parameters and have primarily demonstrated correction of the flow rate parameter inside a single geometry used in both training and testing. Additionally, some machine learning models require a reference object for comparison, which restricts their effectiveness for custom parts \cite{delli2018automated}. Furthermore, these methods are not capable of real-time correction, meaning that if an error is detected, the part cannot be recovered. Although Brion et al. \cite{Brion_Pattinson_2022} addressed this by splitting the toolpath for a layer into smaller segments, optimizing one segment does not necessarily ensure that subsequent segments will also be optimized, especially if their shapes differ significantly.

Moreover, deep learning requires a large labeled dataset, necessitating many prints, which complicates its implementation for error detection due to the high costs and effort needed to generate sufficient training data. Additionally, deep learning models are usually tailored for specific tasks and perform exceptionally well within those domains but often struggle to adapt to different tasks or scenarios. This lack of flexibility, compared to human problem-solving abilities, underscores the challenge of developing a framework that can adapt to various printer setups, detect multiple print defects, and optimize parameters for different parts. Leveraging pre-trained networks could be a potential solution to this problem.

Leveraging transformer architecture \cite{vaswani2017attention,lin2022survey} and massive datasets \cite{brown2020language,chowdhery2023palm,chung2022scaling,chang2023survey,Zhao2023A,achiam2023gpt}, LLMs have made significant strides in various natural language processing (NLP) tasks, including text generation and following task-specific instructions \cite{Zhou2023Instruction-Following, Wang2022Self-Instruct:}. Additionally, LLMs have demonstrated emergent reasoning capabilities by interpolating and utilizing their extensive training data, allowing them to make inferences, draw conclusions, and solve problems beyond their explicit programming \cite{Webb2022Emergent,Huang2022Towards}. Despite the substantial computational resources required for training and fine-tuning LLMs for specific applications, these models have shown an exceptional ability to generalize to new tasks and domains. This remarkable capability is primarily attributed to the in-context learning (ICL) paradigm \cite{brown2020language}. By using minimal natural language prompts and avoiding extensive fine-tuning, LLMs have emerged as effective "few-shot learners" \cite{Lin2021Few-shot, Perez2021True}, demonstrating proficiency with limited training examples.


The adaptability of LLMs to new domains and their ability to learn in context has been transformative across numerous scientific fields. In chemistry, for example, LLMs have autonomously designed, planned, and executed complex experiments \cite{boiko2023autonomous, bran2023chemcrow}. In mathematics and computer science, these models have discovered novel solutions to longstanding problems such as the cap set problem and optimized algorithms for challenges such as the bin-packing problem \cite{romera2024mathematical}. Furthermore, LLMs have significantly advanced research in biomedical fields \cite{thapa2023chatgpt, chen2023extensive, lee2020biobert}, materials science \cite{zaki2023mascqa, xie2023large}, and environmental science \cite{zhu2023chatgpt}. Their impact extends to other scientific domains as well, enhancing understanding and expanding capabilities \cite{holmes2023evaluating, qin2023chatgpt, zeng2023large, wang2023voyager,bartsch2024llm}. Additionally, LLMs have proven to be effective optimizers for foundational problems such as linear regression and the traveling salesman problem, often matching or exceeding the performance of specialized heuristics through straightforward prompting \cite{yang2023large}. This versatility and efficacy underscore the potential of LLMs to drive innovation and efficiency in a wide array of scientific and engineering applications.


In the field of mechanical engineering, fine-tuned LLMs have demonstrated exceptional capabilities in advanced tasks such as knowledge retrieval, hypothesis generation, and agent-based modeling. They have played a crucial role in integrating diverse domains through the use of knowledge graphs \cite{buehler2024mechgpt, chandrasekhar2024amgpt}. Additionally, LLMs excel in various design-related tasks, including sketch similarity analysis, material selection, engineering drawing analysis, CAD generation, and structural optimization \cite{picard2023concept, jadhav2024large}. Their ability to seamlessly connect and enhance these tasks highlights their transformative potential in mechanical engineering. Multi-Modal Vision-language models (VLMs), which combine visual and textual inputs, are now extending these capabilities by supporting parametric CAD sketch generation via CadVLM, which uses sketch images and primitive/text constraints to perform autocompletion, autoconstraint, and conditional image-based CAD sketch generation with high accuracy \cite{wu2024cadvlm}. In manufacturing contexts VLMs are also being used to recognize manufacturing features in CAD models without relying heavily on predetermined geometric rules, by using techniques like few-shot learning, multi-view prompts, and sequential reasoning to generalize across processes such as machining, additive manufacturing, sheet metal forming, mold-making, and casting \cite{khan2025leveraging}.

In additive manufacturing specifically, LLMs and VLMs are fundamentally transforming the 3D printing ecosystem across multiple critical domains. Comprehensive evaluations reveal that VLMs exhibit exceptional versatility throughout the complete engineering design pipeline, from initial conceptualization to final manufacturing and quality inspection, with particularly notable efficacy in design for manufacturing (DfM) evaluations tailored to additive processes \cite{picard2025concept}. This has led to the development of specialized manufacturing-oriented VLMs such as MaViLa, which leverages retrieval-augmented generation (RAG) architectures to integrate domain-specific manufacturing knowledge, achieving demonstrably superior performance compared to general-purpose VLMs across critical manufacturing applications \cite{fan2025mavila}.

\begin{figure}[h!]
  \centering
  
  \includegraphics[width=0.90\textwidth]{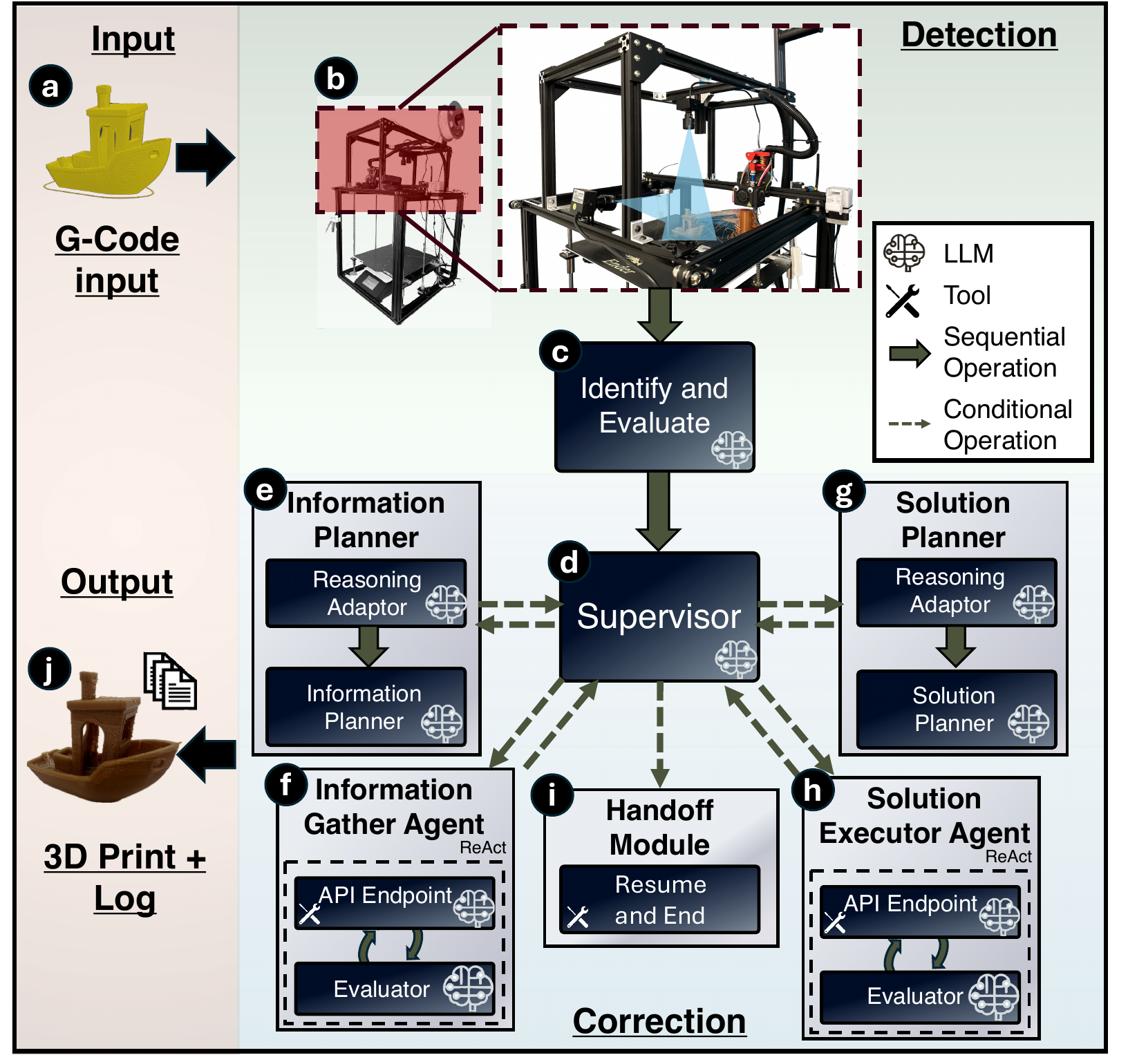}
  \caption{\textbf{Schematic of the Proposed Framework.} (a) The process begins with uploading a part's G-code file to the 3D printer. (b) The printer is equipped with two frame-mounted cameras. (c) After each layer is printed, the extruder moves to the home position, and two images of the current print state are captured. These images are analyzed by the LLM, which evaluates the print, makes observations, and identifies any failures. (d, e) If failures are detected, the LLM supervisor invokes the information planner. (f) The executor then carries out the information gathering plan. (g) Afterwards, the solution planner is activated by the supervisor. (h) The solution plan is executed by another executor. (i, j) Loop ends with the supervisor invoking the handoff module to resume the print.}
  \label{fig:framework}
\end{figure} 

In this study, we introduce a framework that leverages the multimodal capabilities and emergent reasoning capabilities of LLMs to detect and resolve issues during 3D printing. This framework employs specialized LLM agents assigned to specific tasks, coordinated by a supervisory LLM to ensure efficient workflow and communication. By leveraging the strengths of LLMs in reasoning and optimization, the system identifies errors, qualitatively assesses print quality, gathers necessary information, and addresses these issues. This approach allows for the correction of errors in subsequent layers without discarding the entire part, thereby improving efficiency and reducing material waste.

The hierarchical machine-to-machine framework operates by capturing two images of the ongoing 3D print, one from the top and the other from the front, once a layer is completed and the print is paused. These images, along with the natural language description of the part, are fed into an LLM, which evaluates the print quality, identifies defects, and makes relevant observations.

Upon identifying an issue, the supervisory LLM invokes a planner to generate a detailed plan outlining the necessary information and queries for the printer to diagnose the problem. Another LLM agent then executes tools to retrieve this information via the printer’s API. Based on the gathered data, the LLM generates a solution plan, which is implemented by another agent through direct communication with the printer's API. After the solution is executed, the supervisory LLM verifies the parameters and resumes the print. The key advantage of this supervisory LLM is its ability to track the entire conversation among all LLM agents and orchestrate their actions as needed. This comprehensive oversight ensures system coordination and efficiency, with each agent invoked precisely when necessary.

One of the significant benefits of this framework is its flexibility to work across various 3D printers, optimizing print parameters specific to each part without requiring any pre-existing dataset. This adaptability allows the system to fine-tune process parameters dynamically, accommodating different materials, geometries, and printer settings. Additionally, the LLM provides detailed process commentary, aiding in the certification of the part and enhancing traceability. This increases trust in the final product by comprehensively documenting the manufacturing process, reducing the need for destructive testing to validate part integrity. Real-time, insightful commentary on the printing process supports quality assurance and ensures compliance with industry standards and regulations.

\section{Methodology}

\begin{figure}[h!]
  \centering
  
  \includegraphics[width=\textwidth]{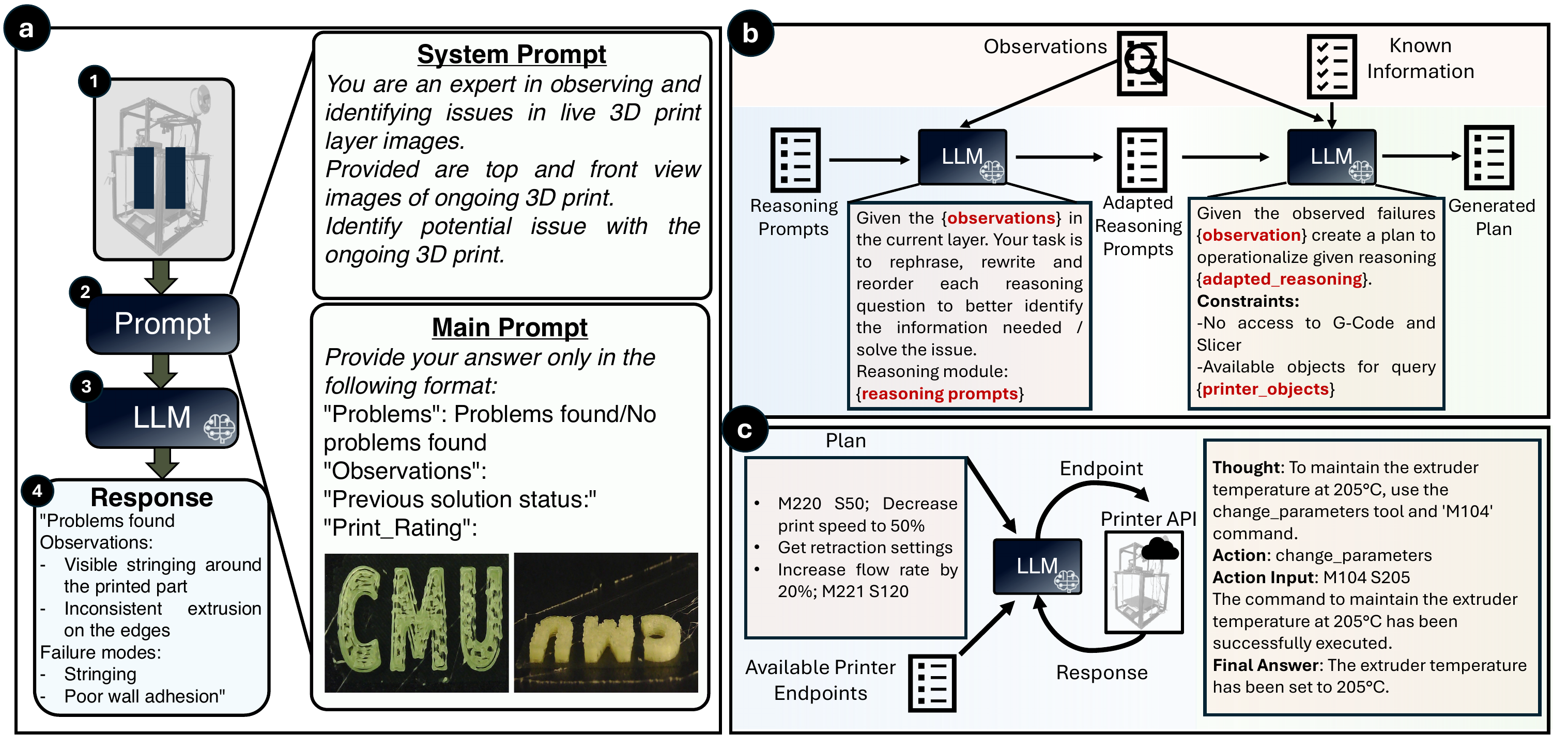}
  \caption{\textbf{Prompt Pipeline for individual agents.} (a) \textbf{Error Detection Agent} – Images from the paused printer (1) are formatted into a (2) predefined prompt and passed to the (3) VLM. The VLM analyzes them to assess print quality and detect failures, returning (4) a structured response. (b) \textbf{Planning Agent} – An adapter module uses an LLM to refine and adapt reasoning prompts based on outputs from the Error Detection Agent. These prompts, together with existing information, guide the planner in formulating an actionable plan. (c) \textbf{Executor Agent} – The LLM selects the appropriate API endpoint to execute the plan on the printer. It validates the command’s success against the printer’s response, adjusting as needed. Internal reasoning and tool use are employed throughout this stage.}
  \label{fig:methods}
\end{figure} 

Given the complexity of the task, our multi-agent LLM framework consists of seven key modules, each designed to perform specific tasks independently as illustrated in Fig.\ref{fig:framework}. The framework includes an Image-Based Reasoning Module (Fig.\ref{fig:framework}c) that identifies defects and makes observations when the printer is paused. There are two planning modules, one focusing on gathering necessary information (Fig.\ref{fig:framework}e) and the other on developing solutions to the identified issues (Fig.\ref{fig:framework}g). Additionally, three execution modules interact directly with the 3D printer to apply these solutions (Fig.\ref{fig:framework}f, \ref{fig:framework}h, \ref{fig:framework}i). A Supervisor Module (Fig.\ref{fig:framework}d) oversees the entire operation, maintaining a dynamic state dictionary that is accessible to other modules, facilitating agent communications, and orchestrating the timely activation of modules based on the current data and needs. By leveraging the specialized abilities and collaborative dynamics among these agents, our multi-agent system not only enhances efficiency in tackling complex tasks but also supports the integration of additional functionalities, allowing for scalable and adaptable enhancements as needed.

\subsection{Error Detection}

Vision-Language Models (VLMs) such as MiniGPT-4 \cite{zhu2023minigpt} and LLaVA \cite{liu2024visual} have demonstrated significant capabilities in understanding and interpreting images, achieving remarkable performance across various visual tasks \cite{Cao_Xu_Sun_Cheng_Du_Gao_Shen_2023,zhu2024llms,elhafsi2023semantic,gu2024anomalygpt}. Owing to extensive pre-training on large datasets of image-text pairs, these models possess exceptional generalization abilities, enabling impressive zero-shot performance. They can effectively recognize and process both familiar and unfamiliar objects without the need for additional training data. This capability to generalize from diverse datasets makes them highly versatile and powerful tools for a wide range of visual and multimodal applications.

  

In this work, we leverage GPT-4o's \cite{achiam2023gpt} image analysis capabilities to detect anomalies in ongoing 3D prints, enhancing the accuracy and reliability of the printing process by identifying and addressing issues in near real-time. After each layer is printed, the printer pauses, and two frame-mounted cameras capture images—one from the top (extruder view) and one from the front—after the extruder returns to its home position. These images along with a description of the part are fed into the LLM for analysis. To prevent redundant error identification, images from the last printed layer are also provided to the LLM, ensuring it does not detect previously addressed issues.

Our methodology employs a finely tuned prompt (SI 1) that instructs the LLM to act as a 3D printing expert agent. The VLM evaluates print quality, identifies visible failure modes, and provides structured, focused, and actionable responses in specified format. 

\subsection{Supervisor Agent}
The Supervisor Agent plays a crucial role in orchestrating the interactions and activities of all modules within the system. It maintains a dynamic state dictionary (SI 4), where agents post updates and outputs from their interactions with the printer. This ensures that each module has access to the most current and relevant information before activation, optimizing the use of LLM tokens for maximum efficiency. As modules complete their assigned tasks, they update their status in the state dictionary. This allows the Supervisor Agent to effectively manage the sequence of module activations, ensuring smooth and efficient transitions between the various stages of the process.


\subsection{Planning Agent}

  

While LLMs have demonstrated emergent reasoning capabilities, their performance can be significantly enhanced when guided by a structured reasoning framework  \cite{Moghaddam2023Boosting, Zhang2023Self-Convinced}. Additionally, because the output of an LLM is heavily influenced by the quality and specificity of the prompts, it becomes essential to adapt these prompts based on the specific problems being addressed. Providing a well-defined reasoning structure and carefully tailored prompts allows the LLM to deliver more accurate and effective solutions, ensuring better outcomes for complex tasks\cite{Significant-Gravitas}.


To achieve this, the planning agent (Fig.~\ref{fig:methods}b) is organized into two modules. The adapter module analyzes observations and detected failures to adjust standard reasoning prompts (SI 2), tailoring them to the specific context while refining and optimizing their relevance. The planner module then leverages these adapted prompts and reasoning frameworks to generate a concrete, actionable plan. Executor agents carry out this plan by interacting with the 3D printer via API to collect data and implement solutions. This dual-module design enables the planning agent to address issues effectively and efficiently during operation.

\subsection{Executor Agent}

  

The Agent Executor (Fig. \ref{fig:methods}c) , a crucial component of the framework, utilizes the ReAct (Reasoning and Acting) method \cite{yao2022react} to execute the generated plans.

The Agent Executor uses a predefined Python function to communicate with the printer via its API. The process starts with the executor receiving a detailed plan from the planning agent, outlining specific actions and corresponding API endpoints. The executor then translates the plan into operational steps by calling these endpoints to run G-code scripts and available macros.

The ReAct method ensures that the executor is not simply issuing commands but actively monitors printer responses and reasons through each step. After executing a command, the executor evaluates the 3D printer's output. If the response is insufficient or indicates an incomplete action, the executor reassesses and adjusts by selecting an alternative endpoint or modifying the G-code script. This iterative approach continues until the desired outcome is achieved, ensuring accurate and efficient execution.

By continuously evaluating API outputs and adapting its actions based on real-time feedback, the Agent Executor effectively handles unforeseen issues and dynamically adjusts its approach.




\section{Experimental Setup}
The printer setup consists of a consumer grade 3D printer modified to stream
in-situ images and system information while also exposing an
Application Program Interface (API) to accept commands to dynamically adjust
printing parameters under the direction of an LLM agent. This is achieved with
Klipper\cite{noauthor_klipper3dklipper_2025}, an open source firmware
project that delegates the task of parsing G-code and process monitoring away
from the controller board to an external computer.
Klipper \cite{noauthor_klipper3dklipper_2025} in conjunction with
ancillary plugins such as
Crowsnest\cite{noauthor_mainsail-crewcrowsnest_2025},
Moonraker\cite{callahan_arksinemoonraker_2025}, and
Mainsail\cite{noauthor_mainsail-crewmainsail_2025} provide a cohesive
interface which allows for monitoring and control over various aspects of the
printing process. 

A stock Creality Ender 5 Plus \cite{creality_3d_technologies_co_ltd_ender-5_nodate} and Creality Ender 3 \cite{creality_3d_technologies_co_ltd_ender-3_nodate} 3D printer
was reflashed with Klipper (v0.13.0)\cite{noauthor_klipper3dklipper_2025} and
configured alongside a standard desktop computer used as the host machine to
deliver commands to the printer's motor control unit. Two SVPRO1080P
cameras with a 2.8 mm to 12 mm range of manual focus were placed to provide top
and front views of the print (Fig. \ref{fig:framework}b). These cameras are
configured with the
Crowsnest (v4.1.16)\cite{noauthor_mainsail-crewcrowsnest_2025} plugin to stream
images of the printing process from the host computer. Of the two cameras, the
top one captures in-situ images of the printing process and these images are
then provided as input to the multimodal LLM to identify and evaluate any
defects in the ongoing print. An input image is captured using with this camera
at the specified end of a sequence of GCode commands. During the capture process the print is paused and toolhead is parked at home positions to avoid toolhead interference over the printed segment. 
These process parameter changes are made using the
Moonraker (v0.9.3)\cite{callahan_arksinemoonraker_2025} API which exposes access
to configurations such as the toolhead temperature, print speed, extrusion rate, or fan speed, to name a few.

Our multi-agent hierarchical framework is implemented in Python (v3.12.4) and builds upon LangChain (v0.3.27) \cite{langchain} along with LangGraph (v0.6.7) \cite{langgraph}. To demonstrate that our framework is robust to shifts in model behavior over time \cite{chen2024chatgpt}, we evaluated it using two large language models: ChatGPT-4o \cite{openai_gpt4o} and ChatGPT-4.1 \cite{openai_gpt41}. This ensures that our results are not tied to one specific model version, but generalize across newer and evolving LLMs.

In our evaluation, we used two different filament materials: PLA and TPU. To introduce environmental variability into the testing setup, the PLA filament had a thermochromic property that changes color from green to brown above 31\degree C, mimicking lighting inconsistencies or thermal gradients often encountered in uncontrolled environments. However, during the printing process, no noticeable color change was observed, suggesting that the thermal conditions remained stable throughout the print. As a result, the color-shifting property was not factored into the defect detection analysis. TPU was used in parallel to test the framework's adaptability to more flexible and mechanically demanding materials.

To ensure robust and reliable interaction with the printer, the LLM was provided with a curated list of modifiable G-code objects derived from a summarized snapshot of the printer's firmware. This approach was necessitated by early observations that supplying the full firmware documentation exceeded the model’s 128k token context limit, occasionally leading to unintended LLM-driven shutdowns—typically when the model misinterpreted persistent errors as indicators of critical system failure. To mitigate this, the curated summary excluded only sensitive or potentially destabilizing commands, such as those related to shutdown, restart, or permanent firmware modification. With these safeguards the LLM retained the ability to freely access and adjust a wide range of operational parameters essential to print quality, including fan speed, bed temperature, cruise speed, acceleration, etc. It could also dynamically query printer state via the API, maintaining its ability to reason over system conditions and implement corrective actions. This careful balance between safety and control enabled effective closed-loop operation without compromising hardware integrity. Additionally, we choose to ignore layer shift defect as such a defect is catastrophic to the print and it cannot be solved without discarding prints.

\section{Result}
We assessed the effectiveness of our framework on both multi-layer and single-layer 3D prints. 

For the multi-layer prints, we conducted two primary experiments: one involving the printing of a wrench and the other featuring raised text, designed to simulate the complexities of multi-part prints. In these experiments, the optimization was done after the completion of each layer, allowing us to continuously monitor and optimize the printing process throughout the entire build.

To assess the effectiveness of the proposed framework at different sampling rates and its ability to identify and correct issues within a single layer, we conducted single-layer tests. These single-layer prints are particularly challenging yet crucial, as they establish the foundation for the entire print. For this analysis, we printed a 100mm x 100mm square with a height of 0.5mm, consisting of a single layer.

The single-layer print was divided into four distinct segments, each representing a different phase of the printing process. After completing each segment, an image was captured and analyzed by the LLM to identify potential failure modes. This analysis focused on detecting inconsistencies and errors through visual observations, providing insights into print quality at each stage. The objective was to determine how well the framework could maintain print integrity and correct issues within the foundational layer, which is critical for the success of the overall print.

Additionally, to evaluate the framework's adaptability to different materials and its ability to optimize parameters, we conducted tests using two different materials: PLA and TPU. Both prints (single layer) were performed at a temperature of 190 \textdegree  C, with the only variation being the print speed, which was set to 120 mm/s, and a layer height of 0.35mm. These tests aimed to verify the framework's capability to work effectively across different materials and optimize parameters accordingly.

\subsection{Multi Layer Prints}

\begin{figure}[h!]
  \centering
  
  \includegraphics[width=\textwidth]{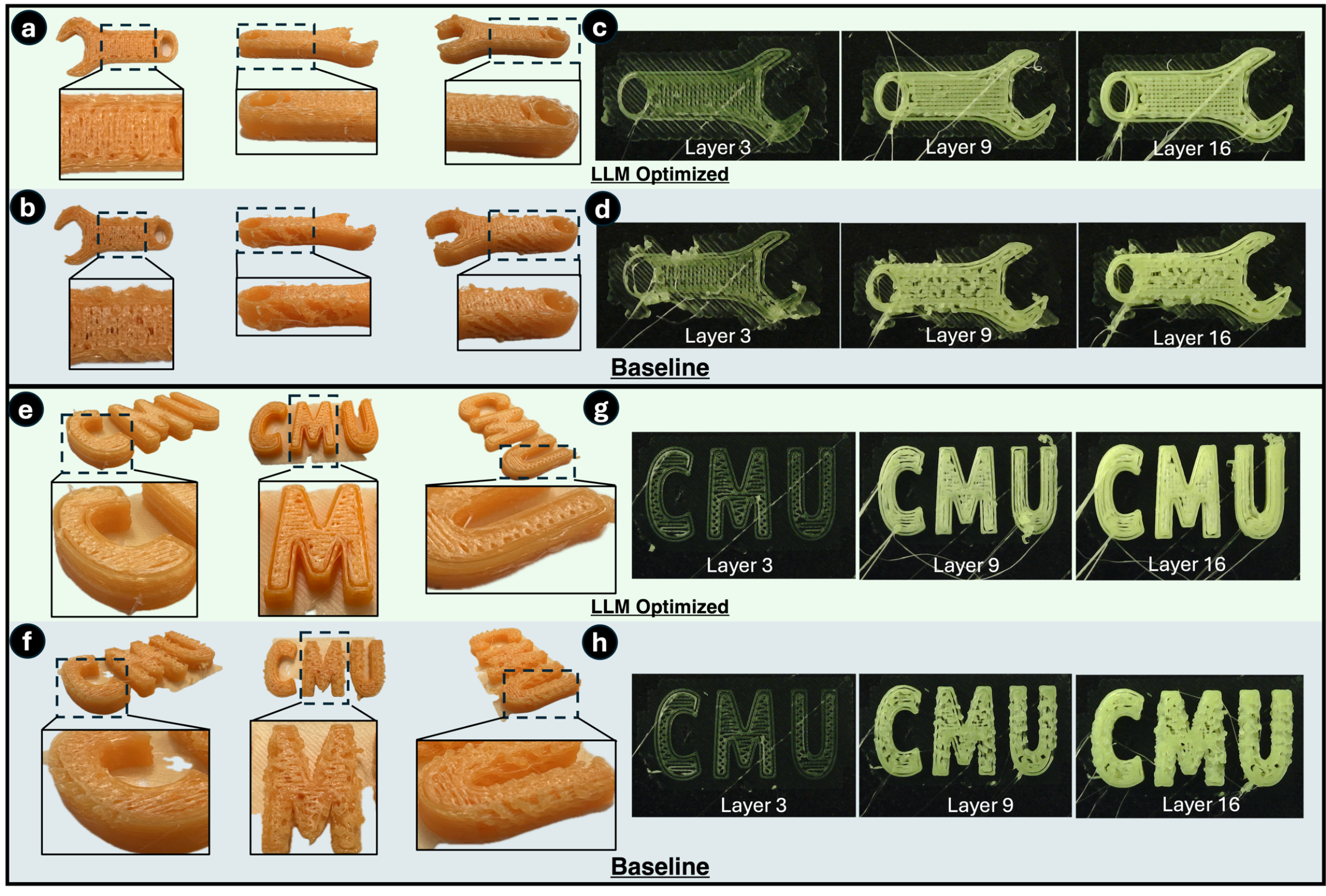}
  \caption{ \textbf{Comparison of LLM-Optimized Print with Baseline:} 
(a) The LLM-optimized print exhibits cleaner, well-defined edges, while the baseline print shows (b) rough, uneven edges.  
(c) The LLM-optimized print maintains a consistently smooth surface finish, whereas the baseline print (d) displays rough surfaces with visible material deposition skips.  
(e) The LLM-optimized print demonstrates consistent extrusion and material deposition, in contrast to the baseline print (f), which suffers from uneven layer adhesion and under-extrusion.  
(g) The LLM-optimized prints show continuous improvements across layers, with better layer adhesion and more precise infill patterns, while the baseline prints display a steady deterioration in print quality.}
  \label{fig:spanner_res}
\end{figure}

To evaluate our framework's ability to detect defects, correct printing errors, and optimize parameters, we used a 3D model of a wrench and raised text, sliced with a rectilinear infill pattern at 100\% infill. Given that the framework requires pausing the print for error analysis and correction—a process that takes time—we added a 20mm diameter cylinder, matching the height of the wrench model, near the extruder's homing position. This cylinder serves to purge oozed material during pauses and prime the nozzle when printing resumes. The toolpath was designed so that, after resuming, the printer first completes the corresponding layer of the cylinder before continuing with the main 3D model, ensuring smooth transitions and consistent print quality.

We selected PLA as the printing material, with the extruder temperature set to 200°C. Recognizing the critical importance of the first layer, default parameters were applied to ensure a solid foundation. For subsequent layers, the extrusion rate was reduced by 25\%, and the print speed was set to 170 mm/s. These settings were chosen to fall within the standard range for PLA, providing a reliable baseline for successful printing while allowing the framework to effectively correct errors and optimize the printing process. 


Additionally, for the print of the wrench model, a manual perturbation was introduced before Layer 9 to disrupt the correct Z-axis movement of the extruder, causing it to shift randomly. This was done to evaluate whether the LLM could accurately detect and respond to sudden, unexpected changes in the printing process. This disruption was not implemented programmatically through the printer's API, as the LLM might have simply traced the last issued commands and recognized any intentional modifications to the Z-offset. By introducing a manual obstruction, the test aimed to challenge the LLM's ability to identify and correct unplanned anomalies.
 
Figure \ref{fig:spanner_res} illustrates a comparison between 3D prints produced using a baseline approach and those optimized by the LLM, highlighting the substantial quality improvements achieved with LLM optimization. In Figure \ref{fig:spanner_res}(a and b), the LLM-optimized print exhibits cleaner, well-defined edges, in contrast to the rough, uneven edges seen in the baseline print Figure \ref{fig:spanner_res}(b and f). The surface finish also shows marked differences, with the LLM-optimized print maintaining a smooth, consistent texture, while the baseline print suffers from rough surfaces and skips in material deposition. Additionally, extrusion consistency is notably improved in the LLM-optimized print, which achieves uniform layer adhesion, whereas the baseline print is plagued by under-extrusion and poor layer bonding. 

Additionally, layer-wise analysis in Fig. \ref{fig:spanner_res}(c and g) reveals continuous improvement in layer quality and infill patterns in the LLM-optimized prints, whereas the baseline prints (Figures \ref{fig:spanner_res}(d and h)) deteriorate as the print progresses. The LLM-optimized approach not only addresses issues such as rough edges, inconsistent surfaces, and poor layer adhesion but also significantly enhances overall print quality, resulting in superior final products.

The stringing issue emerged as a result of the extruder pausing and moving to the home position during the print. The LLM worked to optimize the retraction settings to address this problem. However, despite these efforts, stringing remained unavoidable due to the combination of high flow rate, low print speed, and elevated temperature, which collectively exacerbate the issue. Although the LLM's adjustments provided some improvement, these factors inherently make it challenging to completely eliminate stringing under the given conditions.

  


\subsection{Parameter Optimization}

\begin{figure}[h!]
  \centering
  
  \includegraphics[width=\textwidth]{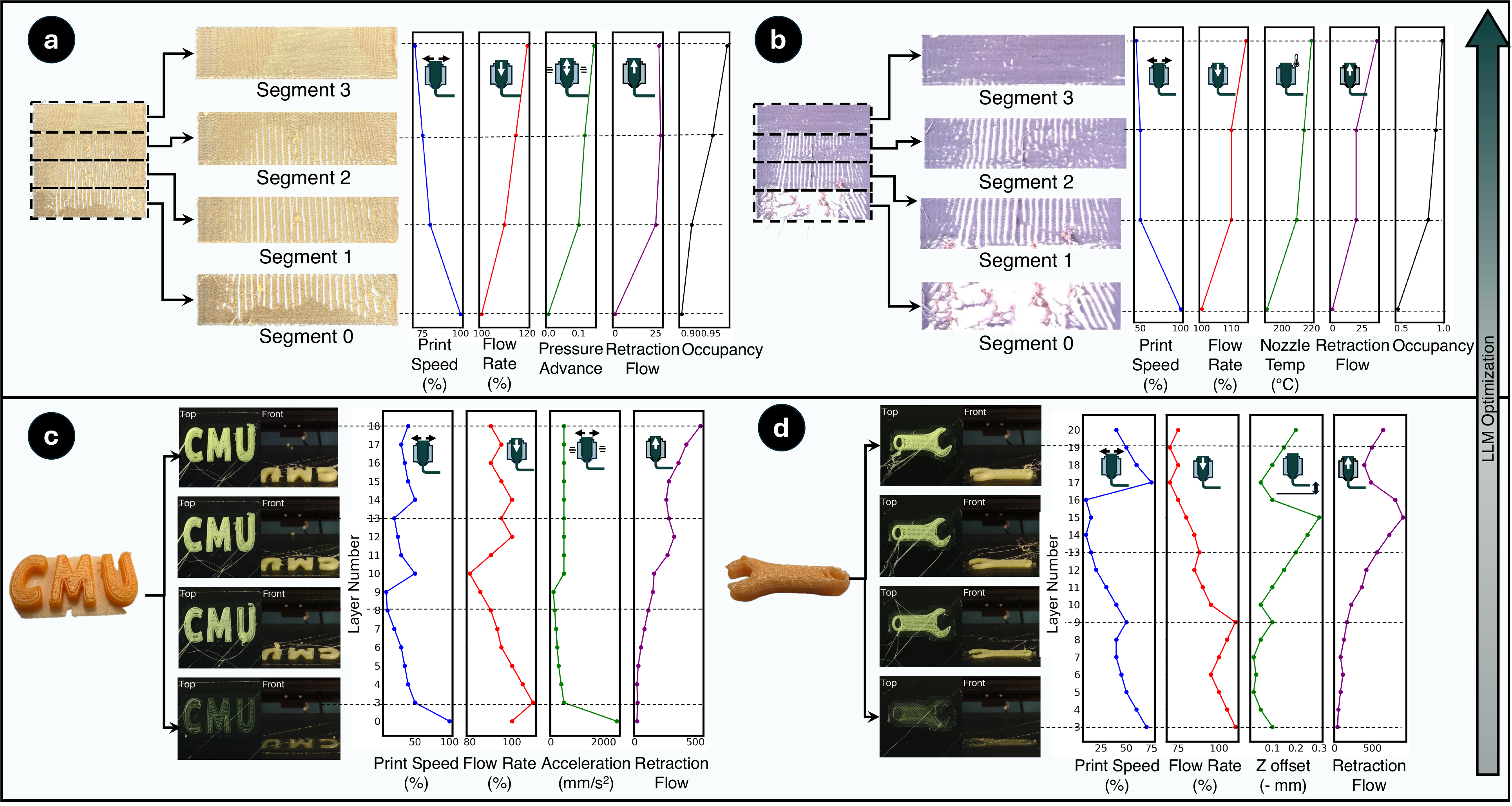}
  \caption{ \textbf{LLM print parameter optimization} For single layer prints using (a) PLA, (b) TPU and for multi-layer prints of (c) text and (d) wrench.}
  \label{fig:optim}
\end{figure} 

To quantify the error/defect in single-layer prints, we define occupancy for each layer as:

$$occupancy = \frac{occupied_{pixels}}{total_{pixels}} $$
This metric provides a measure of how well the printed layer covers the intended area.
For single-layer prints, the LLM identified several key parameters to adjust, including print speed, flow rate, pressure advance, and retraction flow for PLA prints, and nozzle temperature for TPU prints. Fig. \ref{fig:optim} illustrate the changes made by the LLM for single layer, with Fig. \ref{fig:optim}a PLA print and Fig. \ref{fig:optim}b TPU print. For print speed, the LLM suggested reducing it to 75\% to improve filament adhesion and minimize gaps. This change led to smoother deposition and more consistent layers. An adjustment to the flow rate, increasing it slightly above 100\%, addressed the under-extrusion issues observed in the initial segments, resulting in a more uniform filament flow. A minor increase in pressure advance (up to 0.1) helped in managing the flow of filament, especially in corners and intricate details, reducing stringing and blobs. Optimizing retraction flow to around 25\% mitigated the issues related to oozing and stringing between travel moves, enhancing the print's clean lines and precision. For TPU, the LLM suggested raising the nozzle temperature to 220°C, which improved filament melting and flow, leading to better layer adhesion and reduced stringing. The continuous increase in occupancy throughout the single-layer print, as shown, demonstrates the effectiveness of the LLM's optimization strategies in improving print quality.

The ability of the LLM to analyze and adjust parameters for different materials showcases its versatility. The images demonstrate that, following the LLM's recommendations, the quality of the prints improved significantly for both PLA and TPU. For PLA, the adjustments led to smoother layers, better adhesion, and minimized gaps. For TPU, the optimized nozzle temperature and other parameter tweaks resulted in improved layer consistency and reduced stringing, common issues with flexible filaments.

For multi-layer print of wrench as illustrated in Fig. \ref{fig:optim} c, the LLM accurately identifies that the print speed and extrusion rate are off-nominal and actively seeks to optimize these parameters throughout the printing process. Additionally, despite the printer bed being correctly calibrated, the LLM continuously monitors the Z-axis position. Based on the provided layer height and the current layer number, the LLM adjusts the Z-offset of the printer to ensure a smooth surface finish. Notably, at Layer 10, the LLM detected that the Z position was misaligned and persistently attempted to correct it, demonstrating its capability to adapt and fine-tune the printing parameters.

For the print of raised text Fig. \ref{fig:optim}d, the LLM initially identifies signs of under-extrusion and responds by increasing the flow rate, adjusting extruder acceleration, and reducing the print speed. These initial adjustments aim to stabilize the print quality and address the immediate under-extrusion issue. After these corrections, the LLM enters an optimization phase, where it systematically fine-tunes the print speed and flow rate to find the optimal settings for the specific print conditions.

This optimization process proves to be particularly effective in addressing stringing issues, as evidenced by the absence of stringing between the letters in the final print. The LLM's ability to dynamically adjust and optimize key parameters not only enhances the print quality but also demonstrates its capability to mitigate common issues like stringing, which are typically challenging to eliminate in complex prints like raised text.




\subsection{Cross-Platform Validation}
\begin{figure}[h!]
  \centering
  
  \includegraphics[width=\textwidth]{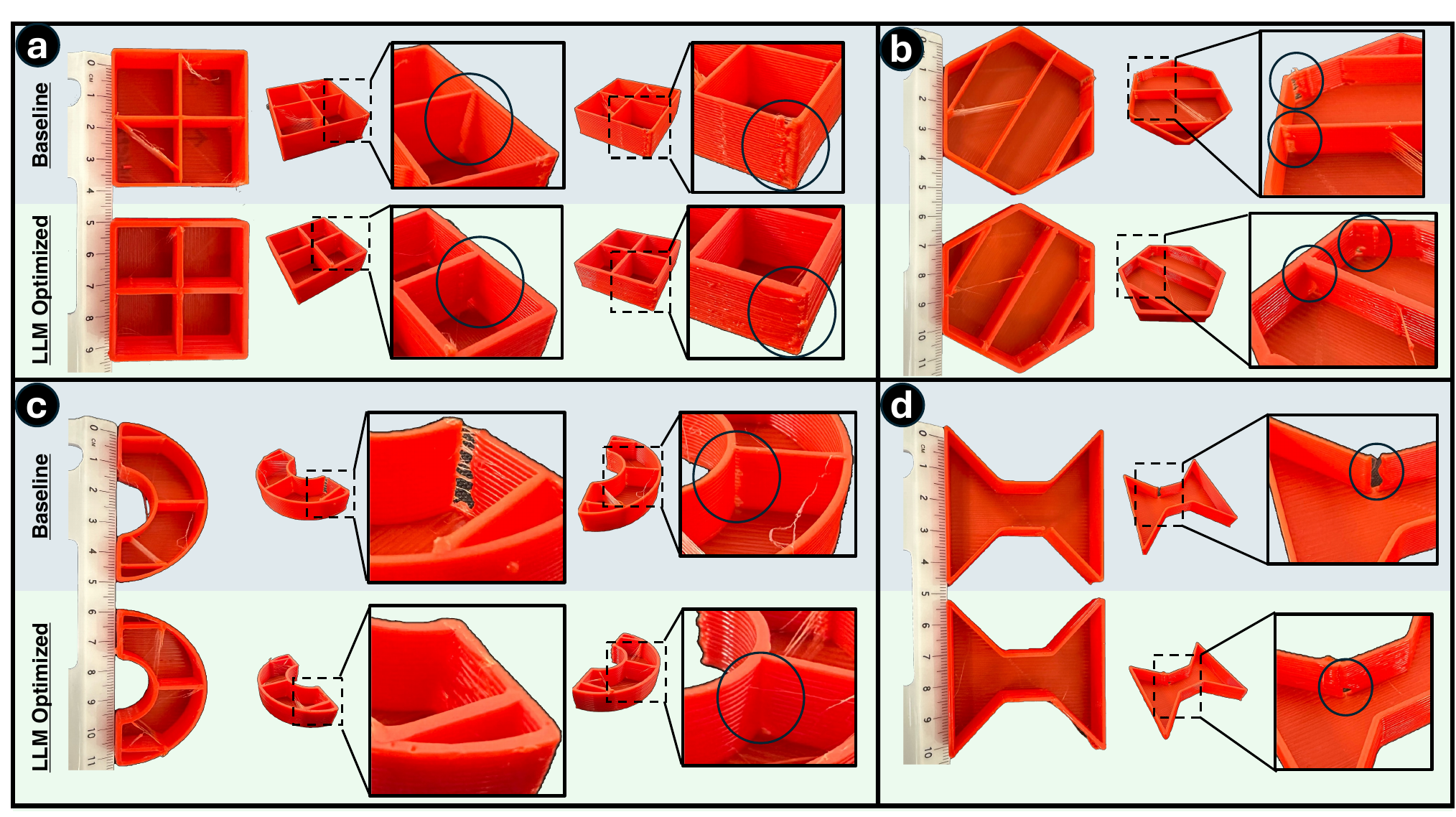}
  \caption{\textbf{Baseline versus LLM-optimized prints.} Baseline samples exhibit visible defects, including seam gaps and incomplete bonding at the infill–perimeter interface, whereas LLM-optimized prints display fully consolidated features across (a) square, (b) hexagonal, (c) hemispherical, and (d) auxetic geometries.}

  \label{fig:llm_strong_quality}
\end{figure} 

To demonstrate the generalizability of our approach across different hardware platforms, we deployed the framework on a Creality Ender 3 printer\cite{creality_3d_technologies_co_ltd_ender-3_nodate}, which features a distinctly different sensor configurations compared to the previously tested Ender 5 Plus\cite{creality_3d_technologies_co_ltd_ender-5_nodate}. Notably, the Ender 3 lacks the BLTouch automatic bed leveling sensor present in the Ender 5 Plus setup. To accommodate this hardware difference, we simply removed the BLTouch-related parameters and queries from the printer documentation accessible to the LLM agents, demonstrating the framework's adaptability to varying hardware configurations. This printer setup utilized the same 1mm nozzle diameter; however, since manufacturer default parameters are typically unavailable for non-standard nozzle sizes, we selected print settings that represented reasonable starting points within the operational envelope, though these settings naturally introduced certain processing challenges that would allow the framework to demonstrate its autonomous error detection and correction capabilities.
Additionally, to evaluate the framework's robustness in addressing defects originating from pre-processing stages, we introduced variations in the infill overlap percentage during the slicing phase, a parameter that becomes embedded within the G-code tool path and remains immutable during runtime execution. This scenario reflects common industrial workflows where G-code files may be sourced from external suppliers or generated using different slicer configurations for intellectual property protection, as manufacturers often receive only the final toolpath data rather than the original design files or slicer parameters. Such situations can result in suboptimal layer adhesion characteristics that cannot be directly corrected through runtime parameter adjustments, thereby testing the framework's ability to compensate for fixed toolpath limitations through optimization of available dynamic parameters such as extrusion rate, temperature, and printing velocities.
Furthermore, to validate that the framework operates independently of visual characteristics such as filament color, we conducted additional prints using PLA material in a different color (RED) from the previous experiments. This color variation test ensures that the image-based defect detection and quality assessment modules function robustly across different visual presentations of the printed material, eliminating potential bias related to specific color properties or contrast characteristics.

Cross-platform validation shows distinct defect patterns in baseline prints, absent in the LLM optimized versions, highlighting the framework’s adaptive correction capabilities. Most prominently, as illustrated in Fig.\ref{fig:llm_strong_quality}, the baseline prints exhibit much better print quality; however, they show consistent gaps at seam positions (the locations where each layer's perimeter and infill path begins and ends), creating potential weak points due to inconsistent material deposition during nozzle acceleration and deceleration. Additionally, noticeable gaps appear at the transition zones where infill patterns connect to perimeter walls, indicating suboptimal overlap parameters embedded in the G-code during slicing. Significantly, the LLM-optimized prints initially display these identical defects in the first few layers, demonstrating that the framework begins with the same compromised conditions. However, within approximately 5 iterations, the system successfully identifies these manifestations as under-extrusion issues and implements corrective measures by adjusting extruder speed, temperature, and extrusion multiplier parameters to compensate for the fixed toolpath limitations. Unlike the previous experiments that began with deliberately poor parameter selections, this validation used near-default printing parameters, resulting in baseline prints with relatively minimal stringing and fewer catastrophic defects, yet the framework still achieved measurable improvements in seam quality, infill-perimeter bonding, and overall surface consistency. This demonstrates the system's sensitivity to subtle quality variations and its ability to optimize even moderately acceptable print conditions toward higher precision manufacturing standards.

\subsection{Experimental Validation}
\begin{figure}[h!]
  \centering
  
  \includegraphics[width=\textwidth]{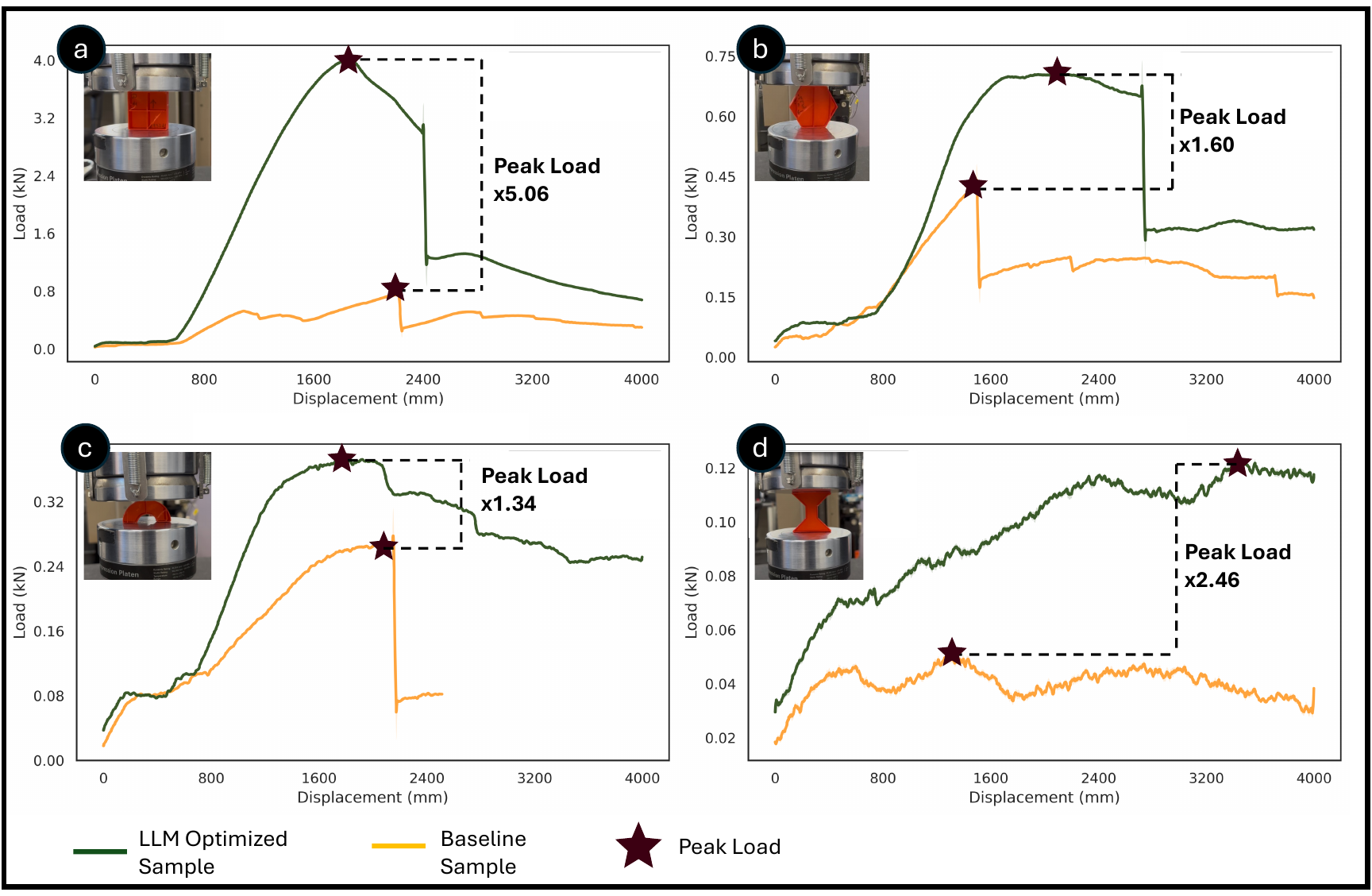}
  \caption{\textbf{Compression performance of baseline and LLM-optimized samples.} Baseline prints show premature failure with lower peak loads, while LLM-optimized counterparts exhibit enhanced structural integrity and significantly higher peak load capacities: (a) square structure (×5.06), (b) hexagonal structure (×1.60), (c) hemispherical structure (×1.34), and (d) auxetic structure (×2.46).}

  \label{fig:llm_compression}
\end{figure}

The observed defects, particularly those occurring at seam positions and infill-perimeter transition zones, represent critical structural vulnerabilities that extend beyond aesthetic concerns. Seam discontinuities create stress concentration points that can initiate crack propagation under mechanical loading, while insufficient infill overlap reduces the effective load transfer between perimeter walls and internal structures. These localized weaknesses can significantly compromise the overall structural integrity of printed components, particularly in applications requiring reliable mechanical performance.
To quantify the mechanical implications of these print quality improvements, we conducted compression testing using an Instron universal testing machine equipped with a 100 kN load cell. Four distinct geometries were evaluated: square structure, hemispherical strucutre, auxetic structure, and a hexagonal unit cell — each representing fundamental building blocks commonly employed in structural applications ranging from packaging to metamaterial design. Given the substantial load capacity of the testing apparatus, the sensitivity limitations for detecting minor load variations necessitated a modification to the LLM optimization strategy. The framework prompts were enhanced to include specific mechanical directives: "strengthen the print in x/y direction for compression test, z is the build direction," ensuring that parameter optimization prioritized structural performance in addition to visual quality.
The compression test results demonstrate substantial mechanical performance improvements across all tested geometries. The square geometry showed the most dramatic enhancement with a 5.06× increase in peak load capacity, rising from approximately 0.8 kN in baseline samples to over 4.0 kN in LLM-optimized prints. The hexagonal unit cell achieved a 1.60× improvement, while the hemisphere and auxetic structures exhibited 1.34× and 2.46× increases, respectively. The load-displacement curves reveal that LLM-optimized samples not only achieve higher peak loads but also demonstrate more consistent mechanical behavior with smoother load progression and more predictable failure patterns. Notably, the baseline samples across all geometries exhibit premature failure and erratic load-displacement relationships, indicating structural inconsistencies that correlate directly with the visual defects identified during the printing process. These results validate that the framework's visual quality improvements translate directly into enhanced mechanical performance, establishing a quantitative link between defect correction and functional part integrity.

\section{Discussion}

\subsection{Analysis of LLM-generated Response}
\begin{figure}[h!]
  \centering
  
  \includegraphics[width=\textwidth]{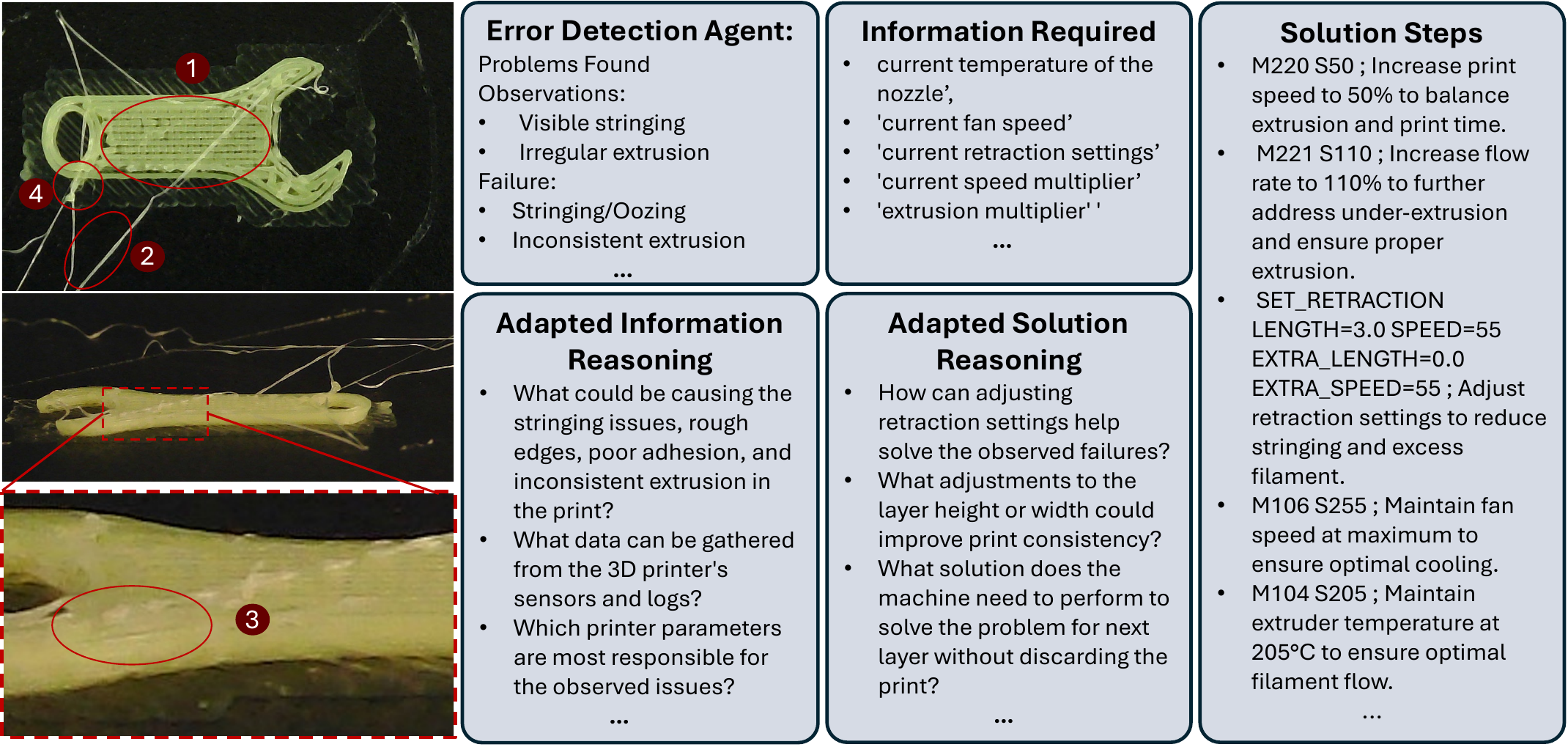}
  \caption{\textbf{LLM-generated response at Layer 9.} Expert-annotated image highlights observed defects: (1) extrusion inconsistency, (2) stringing, (3) layer separation, and (4) material blob. The corresponding LLM-generated response (formatted) identifies these issues and outlines corrective actions based on printer state and visual input.}

  \label{fig:llm_raw}
\end{figure} 

As illustrated in Fig.~\ref{fig:llm_raw}, the expert-annotated illustration identifies four key failure modes: (1) extrusion inconsistency, (2) stringing, (3) layer separation, and (4) a material blob. The LLM successfully identified the first three issues and proposed actionable corrections. However, it did not flag the blob artifact, suggesting some limitations in recognizing less distinct defect types. The blob, formed by excess filament ending in fine strands, visually resembles stringing, which may have led to its misclassification. Importantly, all diagnostic reasoning was performed in-context using a base LLM (ChatGPT-4o) without any retraining or fine-tuning. The reasoning process was guided by domain-specific, generalized structured prompts (e.g., “What could be the reasoning for failure?", "How would you solve the problems without human intervention?”) that we developed in-house.

After identifying the defects, the LLM queried real-time printer parameters such as flow rate, retraction settings, fan speed, and Z-offset, then used this data to produce corrective G-code commands. It increased retraction length and speed to reduce stringing, adjusted the Z-offset for better adhesion, and raised the extrusion multiplier to address under-extrusion. This seamless flow from problem detection to intervention demonstrates the model’s ability to carry out structured, context-aware decisions.

The LLM also considered its previous corrective actions and adapted its strategy. It recognized that earlier adjustments had only partially improved the print and responded with stronger changes, increasing the flow rate from 105\% to 110\% and retraction speed from 50 to 55 mm/s. These changes were not arbitrary. The LLM linked stringing to retraction behavior, attributed poor adhesion to a manually introduced perturbation in the Z-offset at Layer 9, and associated under-extrusion with an insufficient flow rate. This demonstrates the ability to connect causes with the right process variables and propose meaningful adjustments.

By combining live printer data with structured reasoning, the LLM shows potential as an autonomous controller in manufacturing. It does not depend on fixed rules or pre-programmed workflows but instead adapts in real time using prompt-based logic. This makes it a promising tool for scalable and self-correcting quality control in additive manufacturing.

\subsection{LLM Identified Defects}

\begin{figure}[h!]
  \centering
  
  \includegraphics[width=0.9\textwidth]{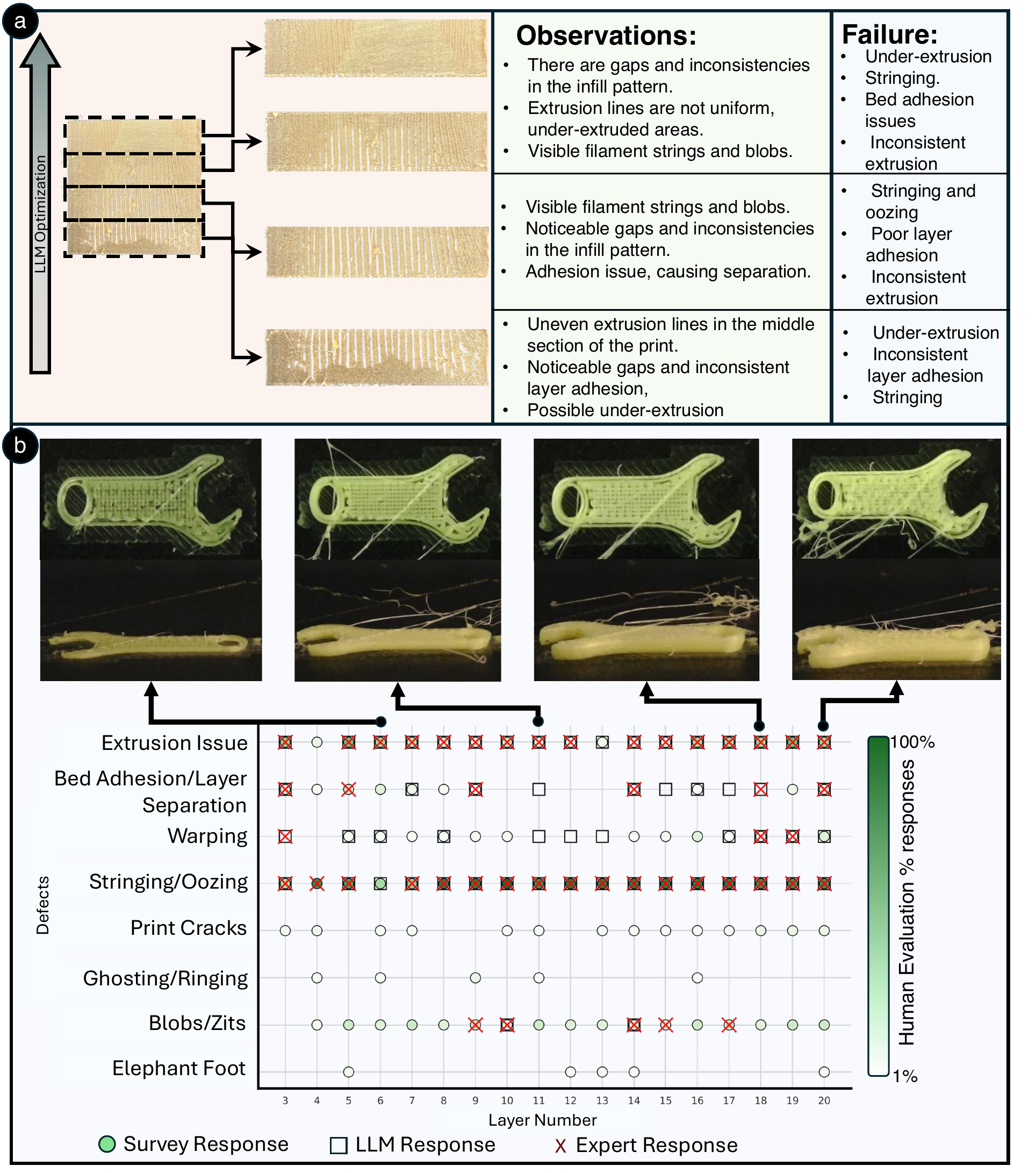}
  \caption{\textbf{Detected Failures.} (a) Failures and observations identified by the LLM during single-layer print optimization. (b) Multi-layer print of the wrench model, showing a layer-by-layer comparison of detected failures across human annotators, domain experts, and the LLM.}

  \label{fig:detect}
\end{figure} 

Error detection is a vital component of the proposed closed-loop framework for additive manufacturing. While LLMs are highly effective at processing textual data and sensor logs, detecting visual or geometric print defects remains a significant challenge, particularly in the absence of fine-tuning on 3D printing-specific datasets. To assess this capability, we evaluated the LLM’s performance in identifying print defects during the fabrication of a single layer, comparing its outputs with expert annotations (SI:4). In the absence of standardized metrics for print quality assessment, such comparisons provide a practical basis for evaluation. Prior work has employed annotated datasets and precision-recall analysis to assess model performance in similar contexts \cite{brion2022generalisable,wang2025transformer}, and we adopt a comparable approach.

In the single-layer prints, the LLM successfully identified several common failure modes, including stringing, under-extrusion, nozzle clogs, inconsistent print speeds, and bed leveling issues. These were inferred from visual indicators such as filament gaps, stray strands, and irregular deposition patterns (Fig.\ref{fig:detect}a). Based on these observations, the model proposed plausible root causes and recommended targeted corrective actions. The model’s outputs showed strong alignment with expert annotations, demonstrating its capability to perform real-time diagnostics on par with domain specialists, even without retraining.

To further evaluate the effectiveness of the LLM (GPT-4o) in detecting errors during multi-layer prints, we conducted a user study involving a diverse cohort of 14 engineers, with varying levels of experience in additive manufacturing. Participants ranged from individuals with basic exposure to AM to those with advanced domain expertise. Each participant was asked to evaluate multi-layer prints that had been previously analyzed and optimized by the LLM. Participants were asked to identify major issues in the ongoing prints, such as poor layer adhesion, surface artifacts, and common defects like under-extrusion, stringing, and warping. Their evaluations were then compared to the LLM's automated detection outputs to assess how effectively the model identified and identified these issues. This comparison enabled us to evaluate not only the LLM’s accuracy in detecting and mitigating print defects, but also how its performance compared to human expertise across varying levels of experience.

Fig.\ref{fig:detect}b presents a comparative analysis of participant responses, expert annotations, and the LLM’s detection results. The analysis shows that both the LLM and the majority of participants, including experts, consistently identified stringing and extrusion-issues as prominent failure modes across multiple layers. In addition, the LLM was able to detect more complex defects, such as bed adhesion failures, warping, and layer separation, which were sometimes overlooked by less experienced participants. These results suggest that the LLM not only matches near-human-level diagnostic capabilities in common error scenarios but can also offer enhanced detection in more subtle or cumulative failure conditions, reinforcing its role as a reliable agent in automated quality monitoring.

It is important to note that some print defects such as cracks, ghosting, ringing, and elephant foot, were reported by a small subset of participants but were not confirmed by expert annotations. This discrepancy likely reflects either over-attribution by less experienced participants or the inherent subjectivity involved in assessing subtle visual anomalies. The relatively low frequency of these responses further suggests that such defects were either inconsistently perceived or misclassified. This divergence between participant feedback and expert assessments reinforces the importance of using expert-labeled annotations as a consistent and reliable ground truth for evaluating the performance of automated systems like the LLM.

\begin{figure}[h!]
  \centering
  
  \includegraphics[width=0.9\textwidth]{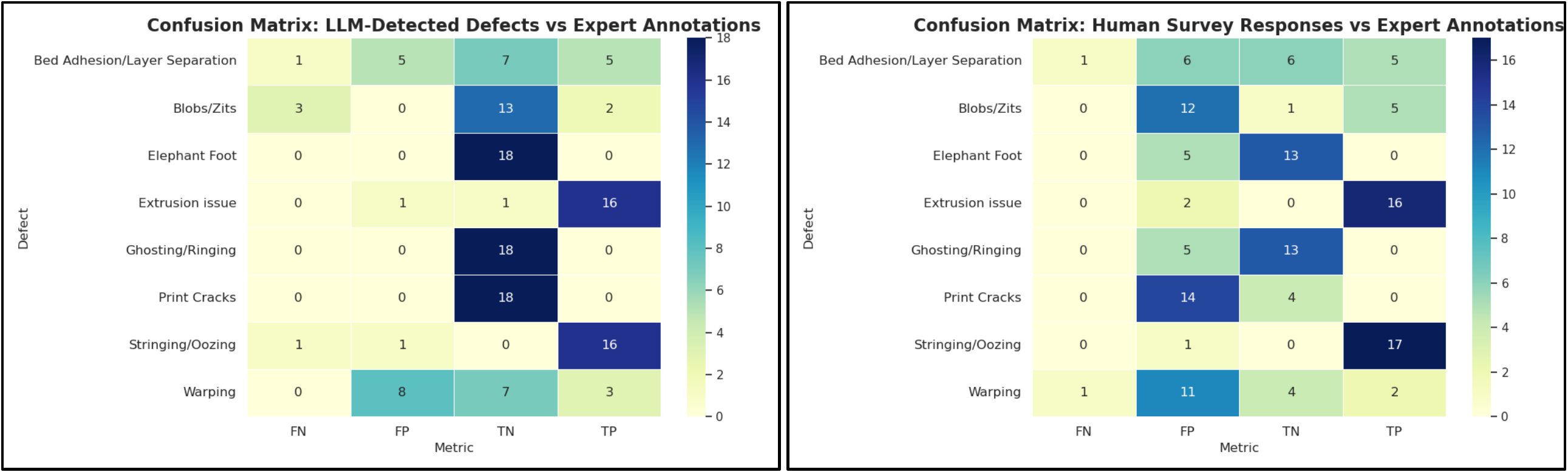}
  \caption{\textbf{Confusion matrix.} Confusion matrices comparing detected print defects against expert annotations. (Left) LLM-predicted defects show high true positive rates for major issues such as extrusion errors and stringing/oozing, but also include a few false positives and misclassifications, particularly for visually similar anomalies like blobs/zits. (Right) Human survey responses show greater variability, with more false positives in subtle defect categories such as print cracks, ghosting, and elephant foot—indicating potential over-identification or misclassification by less experienced participants.}

  \label{fig:confusion}
\end{figure} 

Interestingly, the LLM correctly identified blobs and zits in layers 10 and 14, yet it attributed similar patterns in other layers to stringing or oozing. This suggests that while the LLM demonstrates strong capabilities in recognizing and categorizing major defects, it may sometimes generalize or conflate closely related print anomalies. Such behavior points to both the promise and current limitations of LLMs in fine-grained defect classification, highlighting the need for further refinement to enhance sensitivity to minor or less common print issues.

In some cases, the LLM outperformed human evaluators. For example, it detected early signs of layer separation in layers 7 and 11 before they were identified by expert annotators. This indicates that the model can sometimes anticipate structural failures earlier than humans by reasoning over visual progression across layers.

However, certain defects such as minor warping were occasionally misattributed, and subtle anomalies like layer separation were not always consistently labeled. This may be partly due to limitations in image resolution, as the dataset was compressed to remain compatible with LLM token constraints. Consequently, defects that require high visual fidelity may not be fully captured in the input data, restricting the model’s ability to identify them with high accuracy.

A more quantitative comparison of detection performance is provided in Fig.\ref{fig:confusion}, which presents confusion matrices for both the LLM and human survey responses against expert annotations. The LLM demonstrates strong precision and recall in identifying high-impact defects such as extrusion issues and stringing/oozing, with few false positives or false negatives. In contrast, human survey responses showed greater variability, particularly in the classification of subtle defects such as ghosting, elephant foot, and print cracks. Participants frequently over-identified these rare defects, leading to a higher rate of false positives. This further highlights the LLM’s ability to maintain diagnostic accuracy while minimizing over-attribution.

\subsection{Generalization Without Supervision}
The demonstrated ability to detect and correct 3D printing defects highlights the potential of our approach as a general-purpose solution for process monitoring in additive manufacturing. Unlike conventional deep learning methods that require large annotated datasets and task-specific retraining, our framework leverages prompt-based reasoning and in-context learning to operate without any fine-tuning. The LLM interprets visual cues, queries printer state, and generates corrective actions directly from multimodal inputs. While our evaluation was conducted on a per-layer sampling basis, the framework can be readily extended to operate over multi-layer intervals to reduce the frequency of LLM invocations. Currently, processing a single layer takes between 15 to 45 seconds, depending on the complexity of the failure (judged by number of tokens generated) and whether the model successfully identifies the relevant API endpoints. This latency can be further reduced by explicitly providing the LLM with structured endpoint mappings for each sensor or parameter, improving both response time and reliability.

Recent efforts in automated defect detection have primarily focused on training deep learning models with large-scale annotated datasets, yet these efforts often optimize only for a narrow range of simple failure modes. For instance, one approach trained a convolutional neural network (CNN) on over 120,000 labeled images, using only three coarse defect categories: “Good-quality,” “Under-extrusion,” and “Over-extrusion” \cite{brion2022generalisable}. Similarly, a vision transformer (ViT)-based model relied on heavily curated data and extensive training to identify a limited set of extrusion defects \cite{wang2025transformer}. While such models perform well within their labeled domains, their effectiveness is bounded by the specificity of the annotations. They struggle to generalize to new failure types, printer hardware, or parameter variations not explicitly present in the training data. While the LLM may occasionally misattribute certain defects on a per-layer basis, such as confusing blobs with stringing or overgeneralizing failure categories, its aggregate performance across entire prints leads to significantly improved structural outcomes. By consistently identifying key defect patterns and applying corrective parameter adjustments, the LLM contributes to enhanced print quality over time. 

In contrast, our LLM-based approach leverages general reasoning capabilities and sequential prompting to detect and diagnose a much broader set of failure modes, including stringing, layer separation, warping, and bed adhesion issues. It attributes faults to relevant printer parameters and proposes corrective actions, without relying on domain-specific supervision or pre-labeled examples. This transition from supervised learning to inference-driven reasoning marks a fundamental shift in approach, enabling scalable, interpretable, and adaptable quality control. By removing the dependence on annotated datasets and retraining cycles, our method represents a more flexible and cost-effective pathway for deploying intelligent monitoring in additive manufacturing environments.

A key advantage of our framework is its modular design, which allows seamless integration of different error detection modules without requiring substantial changes to the overall system. While we demonstrate the utility of a LLM for detection, the architecture can accommodate alternative models, such as domain-specific vision networks, heuristic algorithms, or hybrid sensor fusion modules, simply by replacing the error detection component or any other relevant component. This modularity not only promotes extensibility but also future-proofs the system against advances in detection algorithms.

Our framework was deployed on a standard consumer-grade MEX printer with a 1 mm nozzle, lacking manufacturer-provided settings.  The LLM successfully diagnosed and corrected print issues without device-specific tuning. Its reasoning capabilities extended across incomplete inputs and generalized settings. The framework also adapts dynamically to available sensor data, including temperature, retraction settings, print speed, and G-code offsets, based on firmware accessibility. This sensor-agnostic design ensures compatibility with diverse printer models and configurations, supporting wide applicability without requiring hardware-specific integration.

\subsection{Data Security}
Our architecture strictly enforces a separation of roles: the system can only modify runtime parameters during printing, never altering or overwriting the slicer-generated G-code. This safeguard prevents inadvertent or malicious modifications of toolpaths, which past studies have shown can introduce subtle defects or structural compromises. For example, attackers have demonstrated that small edits to G-code may degrade mechanical performance in additive manufacturing\cite{rossel2025security, beckwith2021needle}.

We adopt a modular deployment approach that assigns modules based on their sensitivity and trust requirements. Modules that perform vision-based defect detection and real-time control may be hosted locally using fine-tuned or in-house models; these operate purely on image data and machine state, which contain no proprietary design geometry. General reasoning tasks, such as prompt adaptation, decision planning, and high-level correction logic, can run on external LLM services. Because these reasoning modules consume only abstracted defect descriptors and machine metrics, not original design files or toolpaths, they pose minimal risk of leaking sensitive data.

This partitioning mirrors best practices in secure AI architectures and edge–cloud systems. It allows computation to shift toward external servers for tasks requiring large model capacity, while retaining sensitive processing on-premises to comply with security policies. In highly constrained environments, the framework supports full local deployment so that no external model ever sees proprietary or sensitive data. This flexibility ensures that organizations can balance performance, resource constraints, and confidentiality needs.

\section{Conclusion}
We developed and tested a framework using GPT-4o to detect and address 3D printing defects in near real-time. The LLM effectively identifies major defects like stringing, oozing, layer separation, and inconsistent extrusion, closely matching human evaluations. By dynamically adjusting print parameters based on real-time analysis, the framework significantly enhances print quality and reduces material waste.

A key strength of the LLM framework is its ability to not only detect defects but also to identify the underlying parameters causing these issues. Upon recognizing a defect, the LLM analyzes related print parameters, such as print speed, flow rate, temperature, and retraction settings, to determine their contribution to the problem. It then initiates an optimization process, adjusting these parameters to correct the detected issues in subsequent layers. This proactive approach allows for continuous improvement of the print quality, minimizing the likelihood of defects recurring and enhancing the overall efficiency of the printing process. The agreement-disagreement analysis between the LLM and human evaluators highlights both the strengths and opportunities for further development. The high agreement rates on key defects underscore the LLM’s effectiveness.

An important aspect of this framework is its ability to generate detailed manufacturing commentary and defect detection reports, which significantly enhance part defect traceability and automated documentation. By providing a comprehensive log of the issues encountered and the adjustments made during the printing process, the LLM facilitates better tracking of part quality from start to finish. This automated documentation not only aids in identifying the root causes of defects but also streamlines quality control processes, making it easier to certify parts for use in critical applications and ensuring compliance with industry standards.

Overall, the integration of LLMs into the 3D printing process represents a promising advancement in additive manufacturing. As these models continue to evolve, their ability to detect and correct a broader range of defects will likely improve, making them invaluable tools in achieving higher precision and reliability in AM.





\bibliography{ref}
\newpage
\section{Supplementary Information}
\subsection{SI 1: Image prompt }
\label{sec:image_prompts}
\begin{figure}[h!]
  \centering
  
  \includegraphics[width=\textwidth]{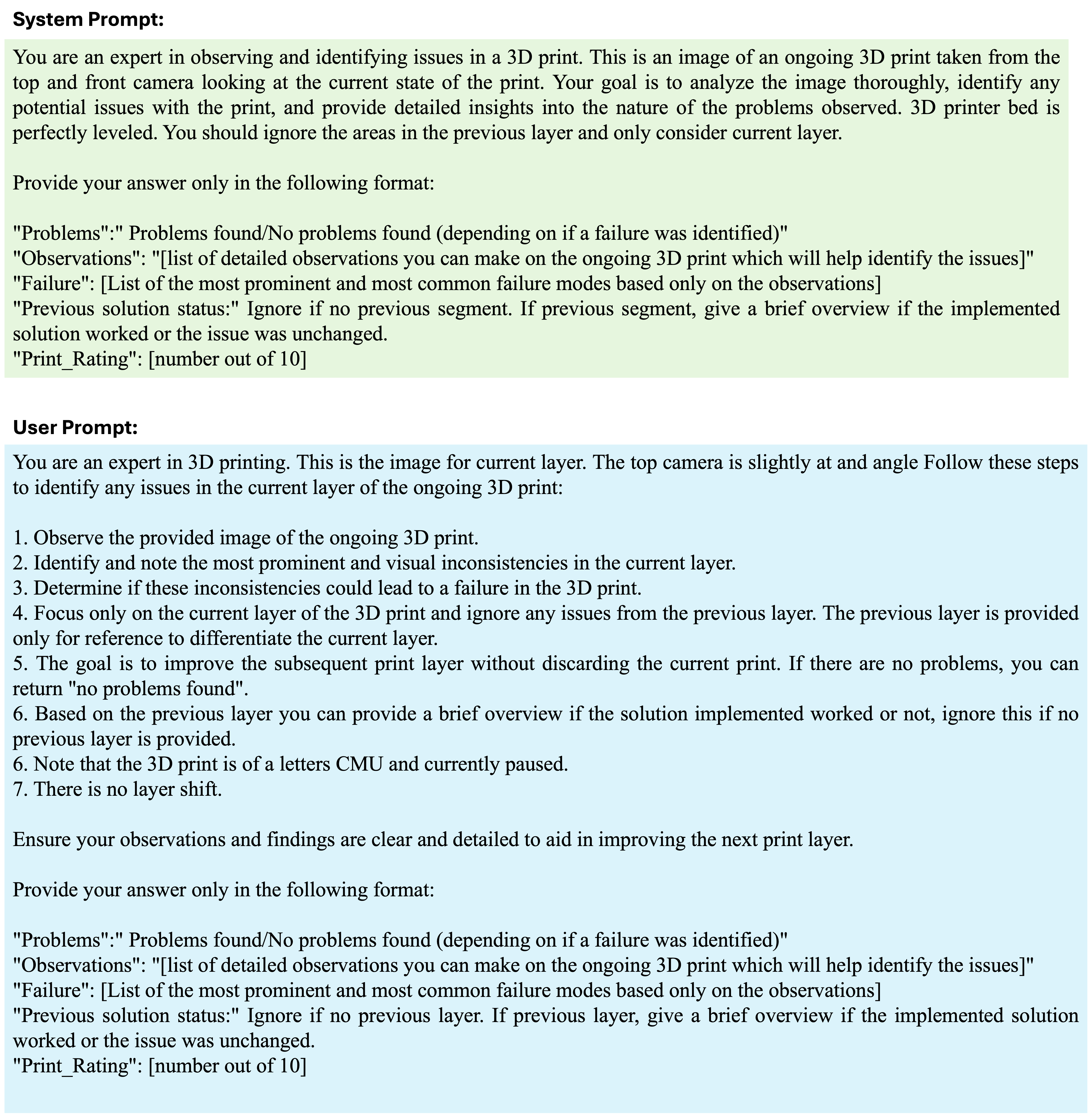}
  \caption*{ \textbf{Image prompt for error detection LLM agent} }
  \label{fig:image_prompt}
\end{figure} 

\newpage

\subsection{SI 2: Reasoning Modules}
\begin{figure}[h!]
  \centering
  
  \includegraphics[width=\textwidth]{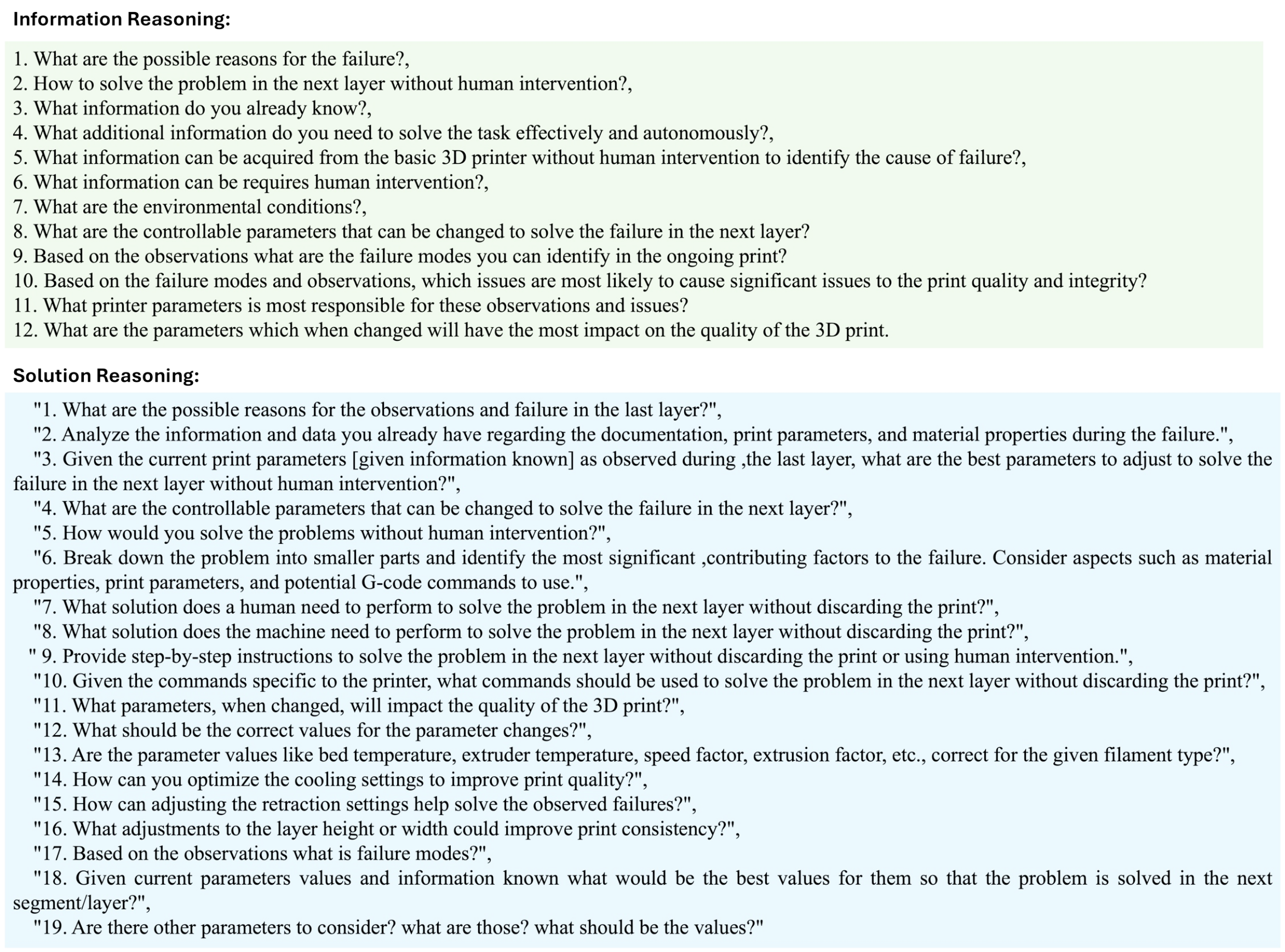}
  \caption*{ \textbf{Reasoning Module for information gathering and solution generation.} }
  \label{fig:reasoning_prompt}
\end{figure} 

\newpage

\subsection{SI 3: Planning agent prompts}
\begin{figure}[h!]
  \centering
  
  \includegraphics[width=\textwidth]{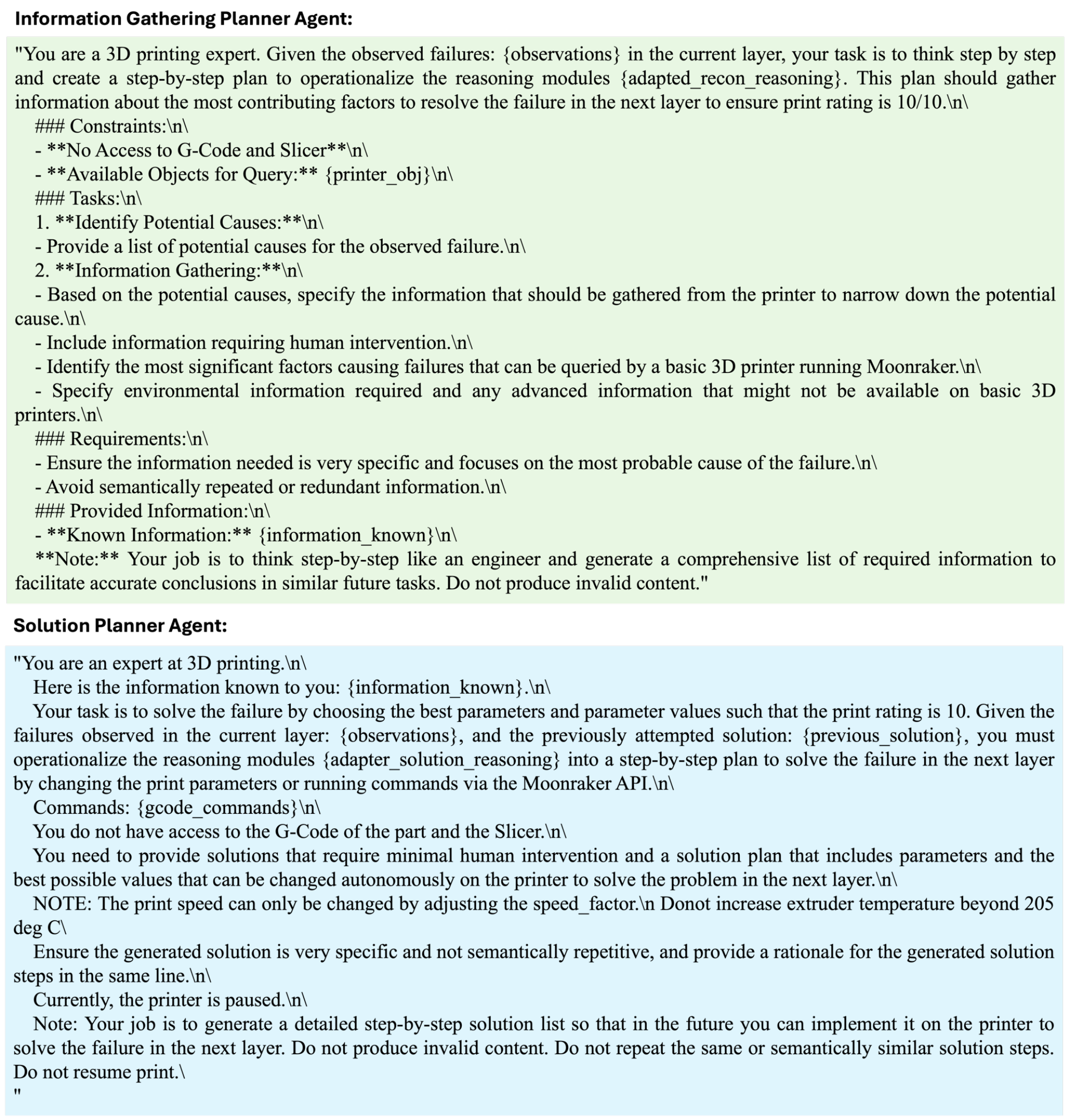}
  \caption*{ \textbf{Prompt for information gathering planner and solution planner.} }
  \label{fig:planning_prompt}
\end{figure} 

\newpage
\subsection{SI 3: Executor agent prompts}
\begin{figure}[h!]
  \centering
  
  \includegraphics[width=\textwidth]{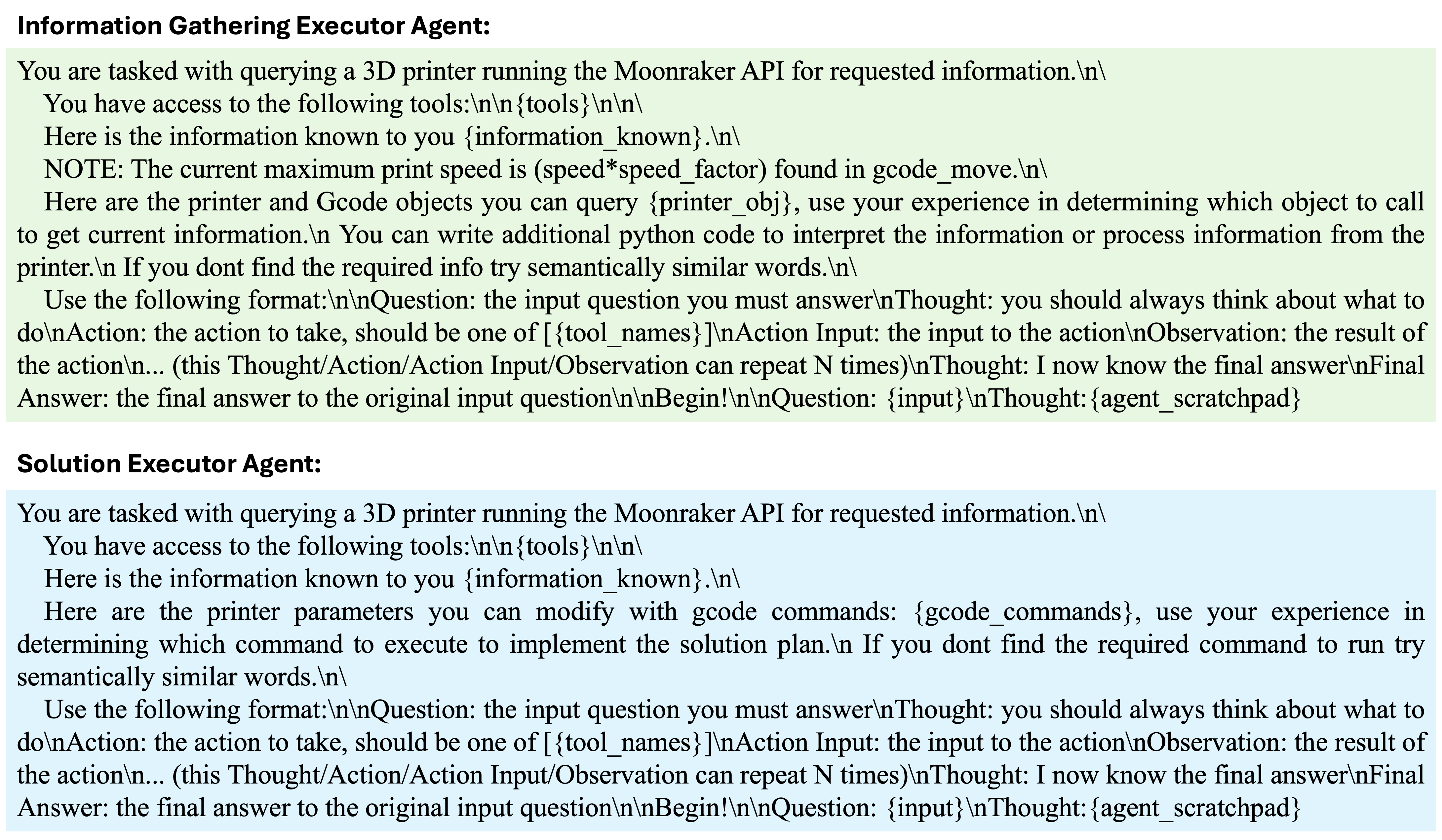}
  \caption*{ \textbf{Prompt for information gathering executor and solution executor.} }
  \label{fig:executor_prompt}
\end{figure} 

\newpage
\section{SI 4: Expert annotations}
\label{sec:annotation}
\includepdf[pages=-]{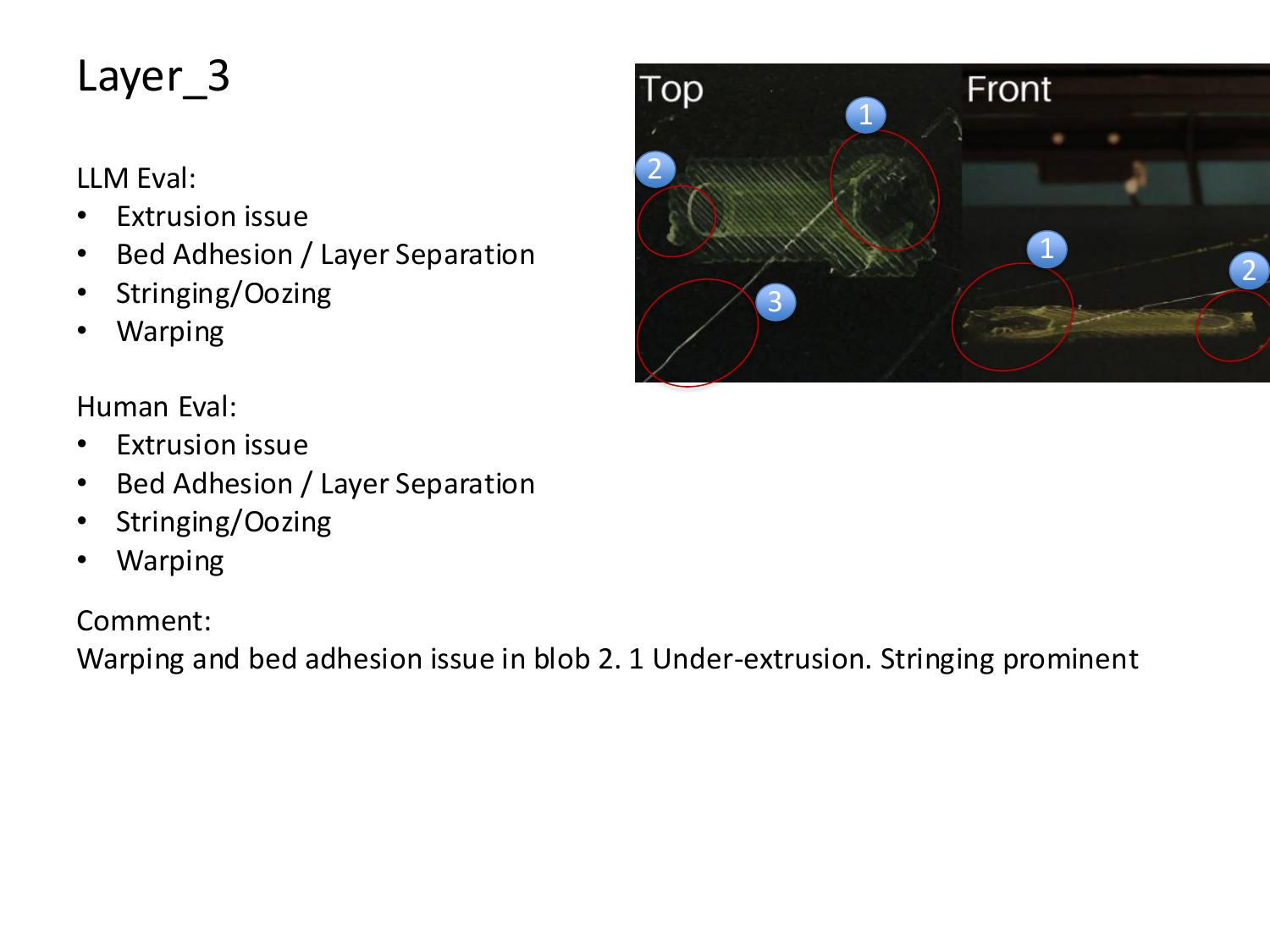}

    
\end{document}